\documentclass[10pt,twocolumn,letterpaper]{article}

\usepackage{cvpr}
\usepackage{times}
\usepackage{epsfig, subfigure, multirow, float}
\usepackage{graphicx}
\usepackage{amsmath}
\usepackage{amssymb}
\usepackage{algorithm, algorithmicx, algpseudocode}
\usepackage[export]{adjustbox}

\usepackage[breaklinks=true,bookmarks=false]{hyperref}

\cvprfinalcopy 

\def\cvprPaperID{2258} 

\ifcvprfinal\fi

\begin{document}

\title{MCMC Shape Sampling for Image Segmentation with Nonparametric Shape Priors}

\author{Ertunc Erdil, Sinan Y{\i}ld{\i}r{\i}m, M\"ujdat \c{C}etin\\
Sabanci University,\\
Tuzla 34956,  Istanbul\\
{\tt\small \{ertuncerdil, sinanyildirim, mcetin\}@sabanciuniv.edu}
\and
Tolga Tasdizen\\
Utah University\\ 
Salt Lake City, UT, USA\\
{\tt\small tolga@sci.utah.edu}
}

\maketitle
\thispagestyle{empty}

\begin{abstract}
Segmenting images of low quality or with missing data is a challenging problem. Integrating statistical prior information about the shapes to be segmented can improve the segmentation results significantly. Most shape-based segmentation algorithms optimize an energy functional and find a point estimate for the object to be segmented. This does not provide a measure of the degree of confidence in that result, neither does it provide a picture of other probable solutions based on the data and the priors. With a statistical view, addressing these issues would involve the problem of characterizing the posterior densities of the shapes of the objects to be segmented. For such characterization, we propose a Markov chain Monte Carlo (MCMC) sampling-based image segmentation algorithm that uses statistical shape priors. In addition to better characterization of the statistical structure of the problem, such an approach would also have the potential to address issues with getting stuck at local optima, suffered by existing shape-based segmentation methods. Our approach is able to characterize the posterior probability density in the space of shapes through its samples, and to return multiple solutions, potentially from different modes of a multimodal probability density, which would be encountered, e.g., in segmenting objects from multiple shape classes. We present promising results on a variety of data sets. We also provide an extension for segmenting shapes of objects with parts that can go through independent shape variations. This extension involves the use of local shape priors on object parts and provides robustness to limitations in shape training data size.
\end{abstract}

\section{Introduction}
Prior knowledge about the shapes to be segmented is required for segmentation of images involving limited and low quality data. In many applications, object shapes come from multiple classes (i.e., the prior shape density is ``multimodal") and the algorithm does not know the class of the object in the scene. For example, in the problem of segmenting objects in a natural scene (e.g., cars, planes, trees, etc.), a segmentation algorithm should contain a training set of objects from different classes. Another example of a multimodal density is the shape density of multiple handwritten digits, e.g., in an optical character segmentation and recognition problem. In this paper, we consider segmentation problems that involve limited and challenging image data together with complex and potentially multimodal shape prior densities.
\begin{figure}[t]
\centering
\begin{tabular}{lllll}
\multicolumn{1}{c}{Test Images} &~& \multicolumn{3}{c}{Samples} \\
 \multicolumn{1}{@{}c@{}}{\includegraphics[width = 1.93cm, height = 1.93cm]{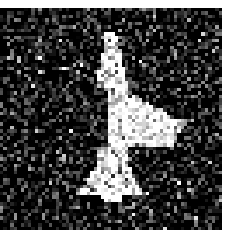}}&~&
 \multicolumn{1}{@{}c@{}}{\includegraphics[width = 1.93cm, height = 1.93cm]{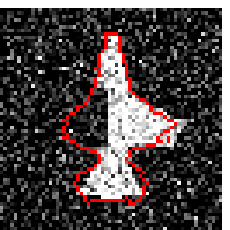}}&
 \multicolumn{1}{@{}c@{}}{\includegraphics[width = 1.93cm, height = 1.93cm]{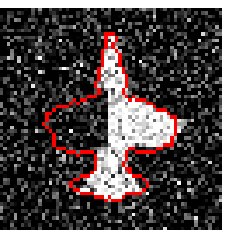}}&
 \multicolumn{1}{@{}c@{}}{\includegraphics[width = 1.93cm, height = 1.93cm]{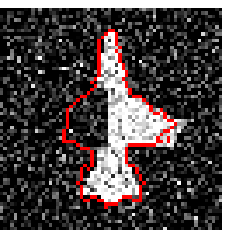}}
\\
 \multicolumn{1}{@{}c@{}}{\includegraphics[width = 1.93cm, height = 1.93cm]{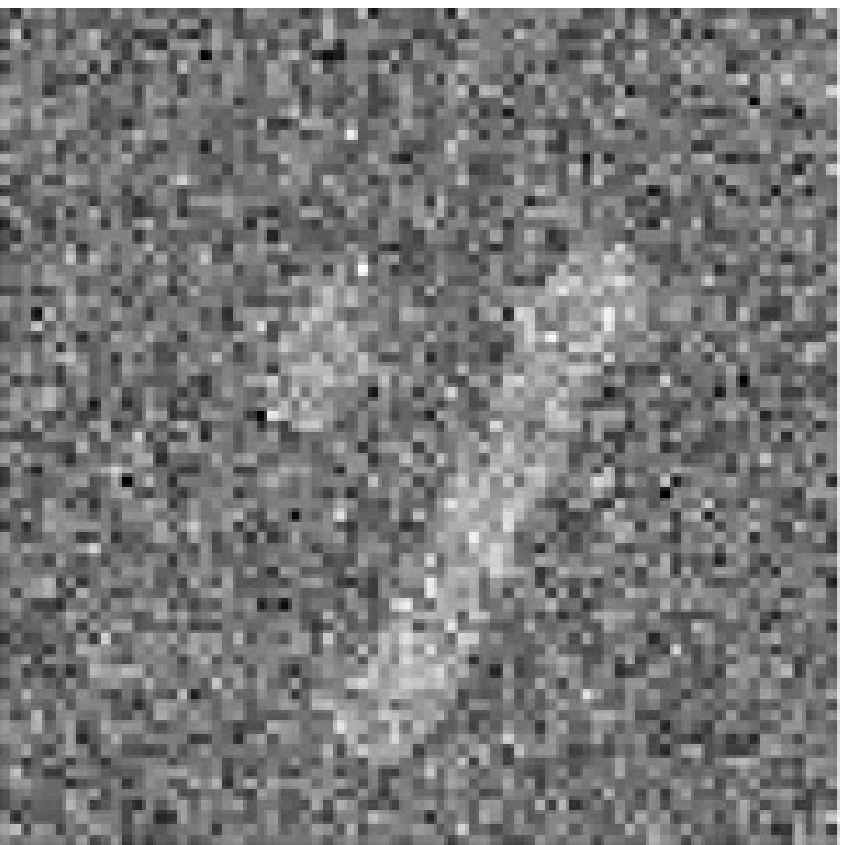}}&~&
 \multicolumn{1}{@{}c@{}}{\includegraphics[width = 1.93cm, height = 1.93cm]{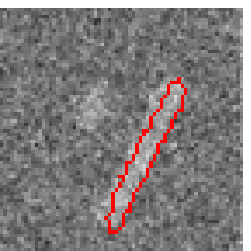}}&
 \multicolumn{1}{@{}c@{}}{\includegraphics[width = 1.93cm, height = 1.93cm]{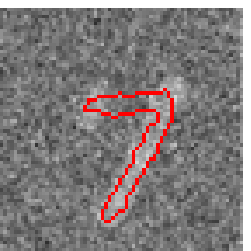}}&
 \multicolumn{1}{@{}c@{}}{\includegraphics[width = 1.93cm, height = 1.93cm]{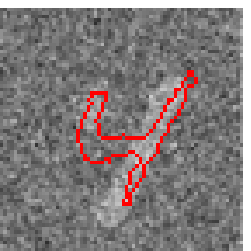}}
 \end{tabular}
 \caption{Examples of MCMC sampling. First row: on an object with a unimodal shape density. Second row: on an object with a multimodal shape density.\label{fig1}}
\end{figure}

The shape-based segmentation approach of Tsai et al. \cite{tsai2003shape} uses a parametric model for an implicit representation of the segmenting curve by applying principal component analysis to a training data set. Such techniques can handle only unimodal, Gaussian-like shape densities and cannot deal with ``multimodal" shape densities. Kim et al.~\cite{kim2007nonparametric} and Cremers et al.~\cite{cremers2006kernel} propose nonparametric methods for handling multimodal shape densities by extending Parzen density estimation to the space of shapes. These methods minimize an energy function containing both data fidelity and shape terms, and find a solution at a local optimum. Having such a point estimate does not provide any measure of the degree of confidence/uncertainty in that result or any information about the characteristics of the posterior density, especially if the prior shape density is multimodal. Such a more detailed characterization might be beneficial for further higher-level inference goals such as object recognition and scene interpretation.

Our contributions in this paper are twofold. First, as the major contribution, we present a Markov chain Monte Carlo (MCMC) sampling approach that uses nonparametric shape priors for image segmentation. Our MCMC sampling approach is able to characterize the posterior shape density by returning multiple probable solutions and avoids the problem of getting stuck at a single local optimum. To the best of our knowledge, this is the first approach that performs MCMC shape sampling-based image segmentation through an energy functional that uses nonparametric shape priors and level sets. We present experimental results on several data sets containing low quality images and occluded objects involving both unimodal and multimodal shape densities. As a second contribution, we provide an extension within our MCMC framework, that involves a local shape prior approach for scenarios in which objects consist of parts that can exhibit independent shape variations. This extension allows learning shapes of object parts independently and then merging them together. This leads to more effective use of potentially limited training data. We demonstrate the effectiveness of this approach on a challenging segmentation problem as well.

Some exemplary results of our MCMC shape sampling approach that uses nonparametric shape priors are illustrated in Figure~\ref{fig1}. The first row of Figure~\ref{fig1} shows three different samples obtained by our approach given the partially occluded aircraft test image in the first column of the corresponding row. In this experiment, the training set contains only examples of aircraft shapes, i.e., the shape density is unimodal, meaning that there are no well-defined subclasses. The second row of the figure contains an MCMC shape sampling example on handwritten digits. In this example, the training set consists of examples from ten different digit classes. Here, our approach is able to find multiple probable solutions from different modes of the shape density. 

\section{Related Work}
Most of the sampling-based segmentation methods in the literature use an energy functional that include only a data fidelity term~\cite{chang2011efficient, chang2012efficient, fan2007mcmc} which means that they are only capable of segmenting objects whose boundaries are mostly visible. Among these approaches, Fan et al.~\cite{fan2007mcmc} have developed a method that utilizes both implicit (level set) and explicit (marker-based) representations of shape. The proposal distribution generates a candidate sample by randomly perturbing a set of marker points selected on the closed curve. Due to the use of marker points in perturbation, this approach is only applicable to segmentation of simply connected shapes, i.e., it cannot handle topological changes. Later, Chang et al.~\cite{chang2011efficient} have proposed an efficient MCMC sampling approach on a level set-based curve representation that can handle topological changes. Random curve perturbation is performed through an addition operator on the level set representation of the curve. Additive perturbation is generated by sampling from a Gaussian distribution. They also introduce some bias to the additive perturbation with the gradient of the energy function to achieve faster convergence. The method is further extended in~\cite{chang2012efficient} to achieve order of magnitude speed up in convergence by developing a sampler whose samples at every iteration are accepted. Additionally, they incorporate topological constraints to exploit prior knowledge of the shape topology.

Chen et al.~\cite{chen2009markov} use the shape prior term suggested by Kim et al.~\cite{kim2007nonparametric} and Cremers et al.~\cite{cremers2006kernel} together with a data fidelity term in the energy functional. Samples are generated by constructing a smooth normal perturbation at a single point on the curve which preserves the signed distance property of the level set. The method is restricted to segmentation of simply connected shapes due to its inability to handle topological changes. Therefore, the approach is not applicable to shapes with complex boundaries.

De Bruijne et al.~\cite{de2004image} propose a particle filter-based segmentation approach that exploits both shape and appearance priors. The method assumes that the underlying shape distribution is unimodal. Therefore, it cannot handle cases when the shapes in the training set comes from a multimodal density.

Eslami et al.~\cite{eslami2014shape} propose a shape model that learns binary shape distributions using a type of deep Boltzmann machine~\cite{salakhutdinov2009deep} and generates samples using block-Gibbs sampling. The model is able to learn multimodal shape densities, however, samples generated from the distribution come only from a particular class closest to the observed data.

\section{Metropolis-Hastings Sampling in the Space of Spaces}
With a Bayesian perspective, segmentation can be viewed as the problem of estimating the boundary $C$ based on image data:
\begin{equation}
p(C|data) \propto \exp(-E(C))
\end{equation}
where,
\begin{footnotesize}
\begin{equation}
E(C) = E_{data}(C) + E_{shape}(C) = -\log{p(data|C)} - \log{p_{C}(C)}
\label{eq:energyFunction}
\end{equation}
\end{footnotesize}

In this paper, we present an algorithm to draw samples from $p(C|data)$ which is, in general, a complex distribution and is not possible to sample from directly.

MCMC methods were developed to draw samples from a probability distribution when direct sampling is non-trivial. We use Metropolis-Hastings sampling \cite{metropolis1953equation} which has been previously used for image segmentation~\cite{fan2007mcmc, chen2009markov, chang2011efficient}. In Metropolis-Hastings sampling, instead of directly sampling from $p$, a proposal distribution $q$ is defined and samples from $q$ are accepted in such a way that samples from $p$ are generated asymptotically. The Metropolis-Hastings acceptance probability is defined as
\begin{footnotesize}
\begin{equation}
\begin{split}
Pr\Big{[}C^{(t+1)} = \mathcal{C}^{(t+1)} | C^{(t)}\Big{]} = \min\Bigg{[} \underbrace{\frac{\pi(\mathcal{C}^{(t+1)})}{\pi(C^{(t)})} .\frac{q(C^{(t)} | \mathcal{C}^{(t+1)})}{q(\mathcal{C}^{(t+1)} | C^{(t)} )}}_\text{Metropolis-Hastings ratio}, 1 \Bigg{]}.
\end{split}
\label{eq:hasting}
\end{equation}
\end{footnotesize}
The Metropolis-Hastings threshold, $\eta$, is randomly generated from the uniform distribution in $[0, 1]$. The candidate (proposed) sample $\mathcal{C}^{(t+1)}$ is accepted if $Pr\Big{[}C^{(t+1)} = \mathcal{C}^{(t+1)} | C^{(t)}\Big{]}$ is greater than $\eta$. Otherwise, $C^{(t + 1)} = C^{(t)}$. In Equation~\ref{eq:hasting}, $C^{(t)}$ and $\mathcal{C}^{(t+1)}$ represent the current sample and proposed sample, respectively. The superscripts $(t)$ and $(t+1)$ denote the sampling iteration count, and $\pi(C) \propto \exp(-E(C))$. After a sufficient number of iterations (i.e., the mixing time) a single sample from the posterior is produced by converging to the stationary distribution. Evaluating the acceptance probability is a key point in MCMC methods. Correct evaluation of the acceptance probability satisfies the sufficient conditions for convergence to the desired posterior distribution: detailed balance and ergodicity. Therefore, the problem turns into the correct computation of forward $q(C^{(t + 1)} | \mathcal{C}^{(t)})$ and reverse $q(\mathcal{C}^{(t)} | C^{(t + 1)})$ transition probabilities of the proposal distribution.

\section{MCMC Shape Sampling using Nonparametric Shape Priors}
We assume that the curve at the $0^{th}$ sampling iteration, $C^{(0)}$, is the curve that is found by minimizing only the data fidelity term, $E_{data}(C)$. We use piecewise-constant version of the Mumford-Shah functional~\cite{mumford1989optimal, chan2001active} for data driven segmentation. One can consider optimizing more sophisticated energy functions such as mutual information~\cite{kim2005nonparametric}, J-Divergence~\cite{houhou2008fast}, and Bhattacharya Distance~\cite{michailovich2007image} to obtain $C^{(0)}$. Also, using an MCMC sampling based approach for data driven segmentation can enrich the sampling space since it would allow subsequent MCMC shape sampling to use several initial curves to start from. After the curve finds all the portions of the object boundary identifiable based on the image data only (e.g., for a high SNR image with an occluded object, one would expect this stage to capture the non-occluded portions of the object reasonably well), we activate the process of generating samples from the underlying space of shapes using nonparametric shape priors.

The overall proposed MCMC shape sampling algorithm is given in Algorithm~\ref{alg1}. The steps of the algorithm are explained in the following three subsections.
\begin{tiny}
\begin{algorithm}
  \caption{MCMC Shape Sampling}
  \label{alg1}
  \begin{algorithmic}[1]
  \For{$i = 1 \to M$}\hfill \Comment{\begin{footnotesize}$M:$ \# of samples to be generated\end{footnotesize}}
  
  	\State Randomly select class of ${C}^{(0)}$ as introduced in Section~\ref{sec:randomClassSelection}.
  	
  	\For{$t = 0 \to (N - 1)$}\hfill \Comment{\begin{footnotesize}$N:$ \# of sampling iterations\end{footnotesize}}
  	
  	\State Generate candidate sample $\tilde{\mathcal{C}}^{(t + 1)}$ from curve $\tilde{C}^{(t)}$ as introduced in Section~\ref{sec:generateCandidateSample}.

\Comment{\begin{footnotesize}The steps between 5 - 10 are introduced in Section~\ref{sec:evaluateMetropolisHastingsRatio}\end{footnotesize}} 
 	
  	\State Calculate Metropolis-Hastings ratio, $Pr$
  	
  	\State $\eta = \mathcal{U}_{[0, 1]}$
  	
  	\If{$(t + 1) = 1$ OR $\eta < Pr$}	
  		\State $\tilde{C}^{(t + 1)} = \tilde{\mathcal{C}}^{(t + 1)}$\Comment{\begin{footnotesize}Accept the candidate\end{footnotesize}}
  	\Else
  		\State $\tilde{C}^{(t + 1)} = \tilde{C}^{(t)}$
  		\Comment{\begin{footnotesize}Reject the candidate\end{footnotesize}}
  	\EndIf
  	
  	\EndFor
  \EndFor
  \end{algorithmic}
\end{algorithm}
\end{tiny}

\subsection{Random class decision}
\label{sec:randomClassSelection}
Suppose that we have a training set $\mathbf{C} = \{C_{1}, \dots, C_{n}\}$  consisting of shapes from $n$ different classes where each class $C_i = \{C_{ij} | j \in [1, m_i] \in  \mathbb{Z}\}$ contains $m_i$ different example shapes. We align training shapes $C_{ij}$ into $\tilde{C}_{ij}$ using the alignment approach presented in Tsai et al.~\cite{tsai2003shape} in order to remove the artifacts due to pose differences such as translation, rotation, and scaling. 

We exploit the shape prior term $p_C(C)$ proposed by Kim et al.~\cite{kim2007nonparametric} to select the class of the curve $\tilde{C}^{(0)}$. The prior probability density function of the curve evaluated at sampling iteration zero is
\begin{equation}
p_{C}(\tilde{C}^{(0)}) = \frac{1}{n} \sum\limits_{i = 1}^{n} \frac{1}{m_i}\sum\limits_{j = 1}^{m_i} k(d_{L_2}(\phi_{\tilde{C}^{(0)}}, \phi_{\tilde{C}_{ij}}), \sigma)
\label{eq:prior}
\end{equation}
where $k(., \sigma)$ is a 1$D$ Gaussian kernel with kernel size $\sigma$, $d_{L_2}(., .)$ is the $L_2$ distance metric and $\phi$ denotes the level set representation of a curve. Also, note that $\tilde{C}^{(0)}$ is the aligned version of $C^{(0)}$ with the training set. By exploiting Equation~\ref{eq:prior}, we can compute the prior probability density of the shapes in $C_i$ evaluated at $\tilde{C}^{(0)}$, $p_{C_i}^{\prime}(\tilde{C}^{(0)})$, as follows
\begin{equation}
p_{C_i}^{\prime}(\tilde{C}^{(0)}) \propto  \frac{1}{m_i}\sum\limits_{j = 1}^{m_i} k(d_{L_2}(\phi_{\tilde{C}^{(0)}}, \phi_{\tilde{C}_{ij}}), \sigma).
\label{eq:propClass}
\end{equation}
We randomly select a class for shape $\tilde{C}^{(0)}$ where the probability of selecting a class is proportional to the value of $p_{C_i}^{\prime}(\tilde{C}^{(0)})$ computed in Equation~\ref{eq:propClass}. When we generate multiple samples, the random class selection step helps us obtain more samples from the classes having higher probabilities.

\subsection{Generating a candidate sample}
\label{sec:generateCandidateSample}
In this section, we explain how to generate a candidate sample from the proposal distribution $q$. Once the class of $\tilde{C}^{(0)}$ is randomly selected, we perform curve perturbation exploiting the training samples in this class. Let $\tilde{C}_{r}$ be the set that contains the training shapes from the selected class $r$. We randomly choose $\gamma$ training shapes from $\tilde{C}_{r}$ where the probability of selecting each shape is proportional to its similarity with $\tilde{C}^{(t)}$. We compute the similarity between a training shape $\tilde{C}_{rj}$ and $\tilde{C}^{(t)}$ as the value of the probability density function, $s$, at $\tilde{C}_{rj}$ where,
\begin{equation}
s_{\tilde{C}^{(t)}}(\tilde{C}_{rj}) \propto k(d_{L_2}(\phi_{\tilde{C}^{(t)}}, \phi_{\tilde{C}_{rj}}), \sigma).
\label{eq:shape_prob}
\end{equation}
Note that a training shape can be selected multiple times and random training shape selection is repeated in each sampling iteration. We represent the set composed of randomly selected $\gamma$ training shapes at sampling iteration $t$ by $\mathbf{\tilde{C}_R}^{(t)}$.

Finally, we define the forward perturbation for the curve $\tilde{C}^{(t)}$ with level sets as follows:
\begin{equation}
\phi_{\tilde{\mathcal{C}}^{(t+1)}} = \phi_{\tilde{C}^{(t)}} + \alpha \mathbf{f^{(t)}}
\end{equation}
We choose $\mathbf{f^{(t)}}$ as the negative gradient of the energy function given in Equation~\ref{eq:energyFunction} in order to move towards a  more probable configuration in each perturbation. Here, $\alpha$ indicates the step size for gradient descent. Note that we use randomly selected training samples, $\tilde{C}_{Rj} \in \mathbf{\tilde{C}_R}^{(t)}$, for curve perturbation. Mathematically this is expressed as
\begin{footnotesize}
\begin{equation}
\begin{split}
&\mathbf{f^{(t)}} = -\frac{\partial E(\phi_{\tilde{C}^{(t)}})}{\phi_{\tilde{C}^{(t)}}} = \frac{\partial \log p(data|\tilde{C}^{(t)})}{\partial t} \\
&+ \frac{1}{p_{\tilde{C}^{(t)}}(\tilde{C}^{(t)})} \frac{1}{\gamma} \frac{1}{\sigma} \sum\limits_{j = 1}^{\gamma} k(d_{L_2}(\phi_{\tilde{C}^{(t)}}, \phi_{\tilde{C}_{Rj}}), \sigma)(\phi_{\tilde{C}_{Rj}} - \phi_{\tilde{C}^{(t)}})
\end{split}
\label{eq:perturbation}
\end{equation}
\end{footnotesize}
In other words, updating the curve $\tilde{C}^{(t)}$ toward the negative gradient direction of the energy functional produces the candidate curve $\tilde{\mathcal{C}}^{(t + 1)}$.

\subsection{Evaluating the Metropolis-Hastings ratio}
\label{sec:evaluateMetropolisHastingsRatio}
Computation of the first fraction in the Metropolis-Hastings ratio in Equation~\ref{eq:hasting} is straightforward since $\pi(C) \propto \exp(-E(C))$. Recall that the candidate curve $\tilde{\mathcal{C}}^{(t+1)}$ is dependent on the forward perturbation $\mathbf{f^{(t)}}$. Therefore, we compute the forward perturbation probability by considering the value of the probability density function, $s$, for each randomly selected training shape $\tilde{C}_{Rj} \in \mathbf{\tilde{C}_R}^{(t)}$ as follows:
\begin{equation}
q(\tilde{\mathcal{C}}^{(t+1)}|C^{(t)}) = \prod \limits_{\tilde{C}_{Rj} \in \mathbf{\tilde{C}_{Rj}}^{(t)}} s(\tilde{C}_{Rj})
\end{equation}

Similarly, the reverse perturbation probability in sampling iteration $(t+1)$ is computed as the probability of selecting random shapes in $\mathbf{\tilde{C}_R}^{(t - 1)}$  which have been used to produce the curve $\tilde{C}^{(t)}$:
\begin{equation}
q(\tilde{C}^{(t)}|\tilde{\mathcal{C}}^{(t+1)}) = \prod \limits_{\tilde{C}_{Rj} \in \mathbf{\tilde{C}_{Rj}}^{(t - 1)}} s(\tilde{C}_{Rj})
\label{eq:reverse}
\end{equation}

Note that, given the above formulations, computation of the reverse perturbation probability is not possible for candidate curve $\tilde{\mathcal{C}}^{(1)}$, the curve at sampling iteration $1$, since we have to use information from sampling iteration $-1$ for evaluation of Equation~\ref{eq:reverse}, which is not available. Therefore, we accept the candidate sample $\tilde{\mathcal{C}}^{(1)}$ without evaluating the Metropolis-Hastings ratio and consider the above-mentioned steps for generating samples after sampling iteration $1$.

\subsection{Discussion on sufficient conditions for MCMC sampling}
\label{sec:sufficient}
Convergence to the correct stationary distribution is crucial in MCMC methods. Convergence is guaranteed with two sufficient conditions: (1) that the chain is ergodic, and (2) that detailed balance is satisfied in each sampling iteration. Ergodicity is satisfied when the Markov chain is aperiodic and irreducible. Aperiodicity of a complicated Markov chain is a property that is hard to prove as attested in the literature~\cite{gilks1996}.

Detailed balance is satisfied as long as the Metropolis-Hastings ratio in Equation~\ref{eq:hasting} is calculated correctly. We have already described how we compute the Metropolis-Hastings ratio in the previous section. Empirical results show that a stationary distribution is most likely reached since our samples converge. Related pieces of work in~\cite{fan2007mcmc},~\cite{chang2011efficient}, and~\cite{chen2009markov} argue that the Markov chain is unlikely to be periodic because the space of segmentations is so large. Similarly, we can also assert that our Markov chain is unlikely to be periodic. Even if the chain is periodic in exceptional cases, the average sample path converges to the stationary distribution as long as the chain is irreducible. Irreducibility of a Markov chain requires showing that transitioning from any state to any other state has finite probability. Chen et al.~\cite{chen2009markov} and Chang et al.~\cite{chang2011efficient} provide valid arguments that the Markov chain is irreducible whereas Fan et al.~\cite{fan2007mcmc} does not discuss this property. As explained in the previous section, curve perturbation in our framework is performed with randomly selected training samples $\mathbf{\tilde{C}_R}^{(t)}$ and each shape has finite probability to be selected at any sampling iteration. With this perspective, we can also argue that each move between shapes has finite probability in our approach. 

\section{Extension to MCMC Sampling using Local Shape Priors}
\begin{figure}[ht]
\centering
\subfigure{
\begin{tabular}{clcc}
\multicolumn{2}{c}{\multirow{2}{*}{\begin{tabular}[c]{@{}c@{}}Training images\end{tabular}}} & \multirow{2}{*}{Test image} & \multirow{2}{*}{\begin{tabular}[c]{@{}c@{}}Segmentation with\\ global priors\end{tabular}} \\
\multicolumn{2}{c}{}                                                                           &                             &                                                                                              \\
\multicolumn{1}{c}{\includegraphics[width = 1.2cm, height = 2cm]{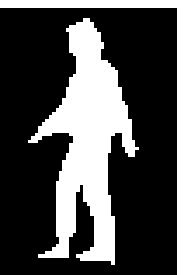}}                                    &\includegraphics[width = 1.2cm, height = 2cm]{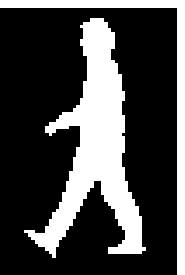}                                   & \multicolumn{1}{c}{\includegraphics[width = 1.2cm, height = 2cm]{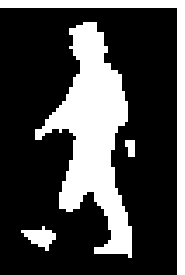}}       & \multicolumn{1}{c}{\includegraphics[width = 1.2cm, height = 2cm]{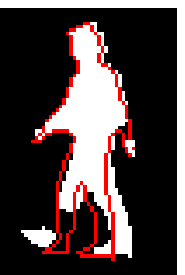}}                                                                       \\
\\
\multicolumn{2}{c}{\multirow{2}{*}{\begin{tabular}[c]{@{}c@{}}Local shape priors\\ with colored patches\end{tabular}}} & \multirow{2}{*}{\begin{tabular}[c]{@{}c@{}}Activated local\\ shape priors\end{tabular}} & \multirow{2}{*}{\begin{tabular}[c]{@{}c@{}}Expected \\segmentation\end{tabular}} \\
\multicolumn{2}{c}{}                                                                           &                             &                                                                                              \\
\multicolumn{1}{c}{\includegraphics[width = 1.2cm, height = 2cm]{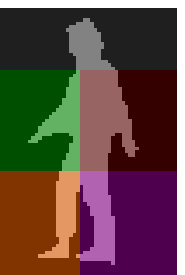}}                                    &\includegraphics[width = 1.2cm, height = 2cm]{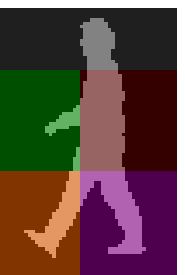}                                   & \multicolumn{1}{c}{\includegraphics[width = 1.2cm, height = 2cm]{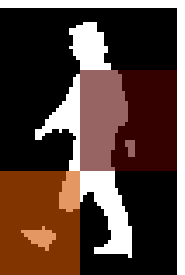}}       & \multicolumn{1}{c}{\includegraphics[width = 1.2cm, height = 2cm]{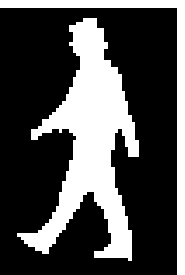}}                                                                      
\end{tabular} 
 }
 \caption{Motivating example for using local shape priors in walking silhouettes data set.~\label{fig:extension_motivation}}
\end{figure}

In this section, we consider the problem of segmenting objects with parts that can go through independent shape variations. We propose to use local shape priors on object parts to provide robustness to limitations in shape training size~\cite{chen2013deep, mesadi2015disjunctive}. Let us consider the motivating example shown in Figure~\ref{fig:extension_motivation}. In this example, there are three images of walking silhouettes: two for training and one for testing. Note that the left leg together with the right arm of the test silhouette involves missing regions. When segmenting the test image using nonparametric shape priors~\cite{kim2007nonparametric} based on global training shapes\footnote{Unless otherwise stated, the shape priors we use are global. We explicitly refer to global shape priors when we need to distinguish them from local shape priors.}, the result may not be satisfactory (see the rightmost image in the first row of Figure~\ref{fig:extension_motivation}), because the shapes in the training set do not closely resemble the test image. This motivates us to represent shapes with local priors such that resulting segmentation will mix and match information from independent object parts (e.g., by taking information about the the right arm from the first training shape and about the left leg from the second training shape). 

Our idea of constructing local shape priors is straightforward. Once the training shapes are aligned, we divide the shapes into patches, such that each patch contains a different local shape region. Each patch is indicated by a different color in the second row of Figure~\ref{fig:extension_motivation}. Note that the patches representing the same local shape have identical size. For MCMC shape sampling using local shape priors, it is straightforward to adapt the formulation in the previous sections to consider local priors. In particular, instead of choosing random global shapes using the values computed by Equation~\ref{eq:shape_prob}, we compute these values for each patch (local shape) and select random patches among all training images. Note that evaluation of forward and reverse perturbation probabilities should also be modified accordingly. 

\begin{figure*}[t]
\centering
\subfigure{
 \includegraphics[width = 1.2cm, height = 1.2cm]{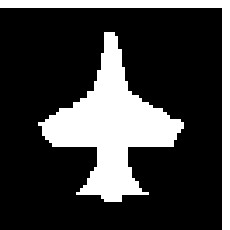}
 \includegraphics[width = 1.2cm, height = 1.2cm]{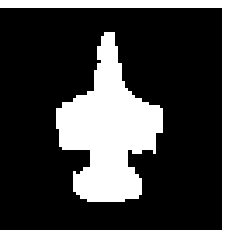}
 \includegraphics[width = 1.2cm, height = 1.2cm]{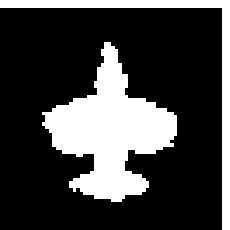}
 \includegraphics[width = 1.2cm, height = 1.2cm]{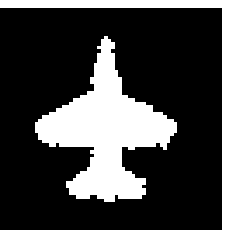}
 \includegraphics[width = 1.2cm, height = 1.2cm]{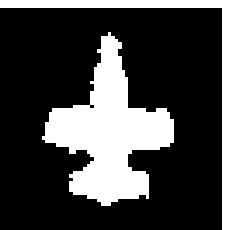}
 \includegraphics[width = 1.2cm, height = 1.2cm]{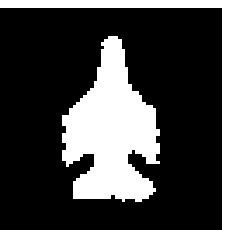}
 \includegraphics[width = 1.2cm, height = 1.2cm]{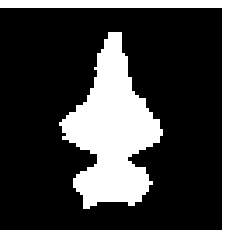}
 \includegraphics[width = 1.2cm, height = 1.2cm]{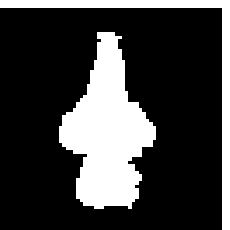}
 \includegraphics[width = 1.2cm, height = 1.2cm]{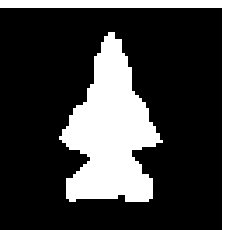}
 \includegraphics[width = 1.2cm, height = 1.2cm]{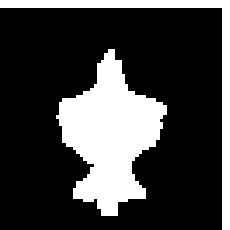}
 \includegraphics[width = 1.2cm, height = 1.2cm]{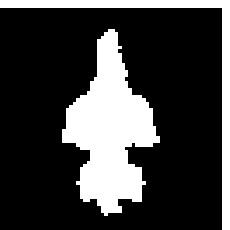}
 }
 \subfigure{
 \includegraphics[width = 1.2cm, height = 1.2cm]{figures/aircraft_testImage1_less_noisy.eps}
 \includegraphics[width = 1.2cm, height = 1.2cm]{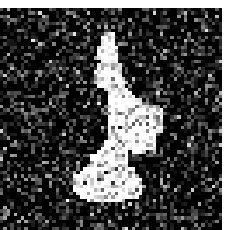}
 \includegraphics[width = 1.2cm, height = 1.2cm]{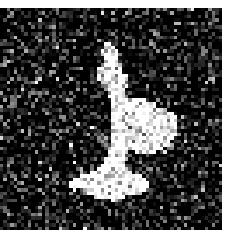}
 \includegraphics[width = 1.2cm, height = 1.2cm]{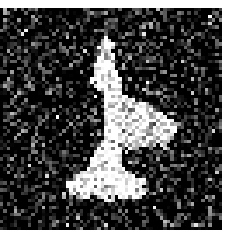}
 \includegraphics[width = 1.2cm, height = 1.2cm]{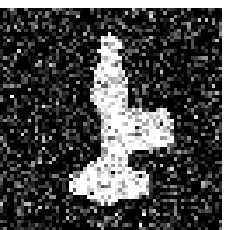}
 \includegraphics[width = 1.2cm, height = 1.2cm]{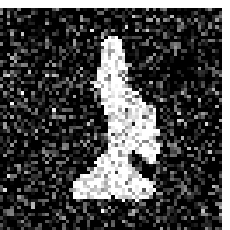}
 \includegraphics[width = 1.2cm, height = 1.2cm]{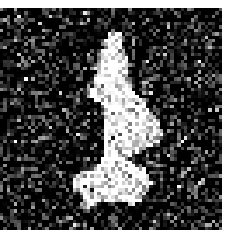}
 \includegraphics[width = 1.2cm, height = 1.2cm]{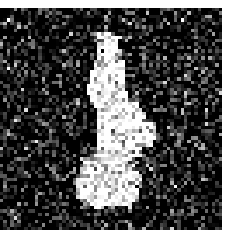}
 \includegraphics[width = 1.2cm, height = 1.2cm]{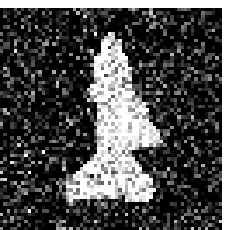}
 \includegraphics[width = 1.2cm, height = 1.2cm]{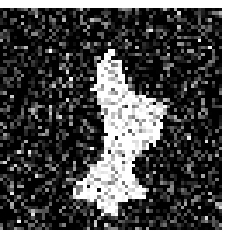}
 \includegraphics[width = 1.2cm, height = 1.2cm]{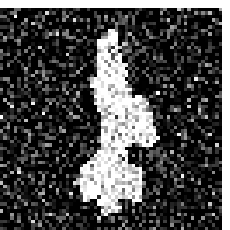}
 }
 \subfigure{
 \includegraphics[width = 1.2cm, height = 1.2cm]{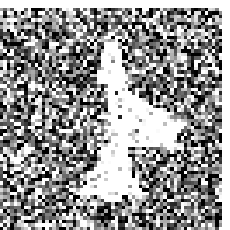}
 \includegraphics[width = 1.2cm, height = 1.2cm]{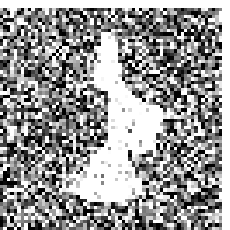}
 \includegraphics[width = 1.2cm, height = 1.2cm]{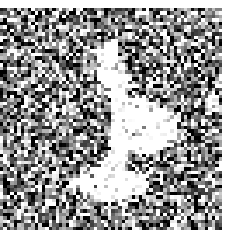}
 \includegraphics[width = 1.2cm, height = 1.2cm]{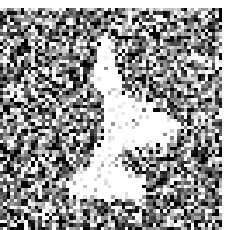}
 \includegraphics[width = 1.2cm, height = 1.2cm]{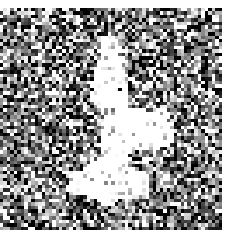}
 \includegraphics[width = 1.2cm, height = 1.2cm]{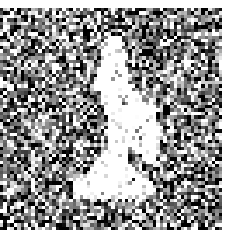}
 \includegraphics[width = 1.2cm, height = 1.2cm]{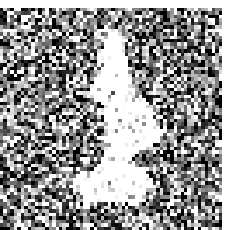}
 \includegraphics[width = 1.2cm, height = 1.2cm]{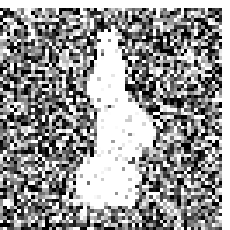}
 \includegraphics[width = 1.2cm, height = 1.2cm]{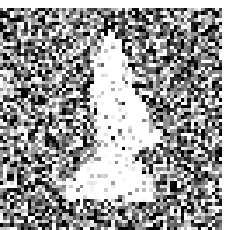}
 \includegraphics[width = 1.2cm, height = 1.2cm]{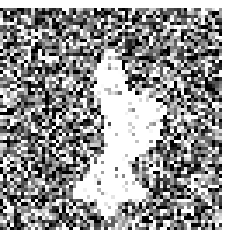}
 \includegraphics[width = 1.2cm, height = 1.2cm]{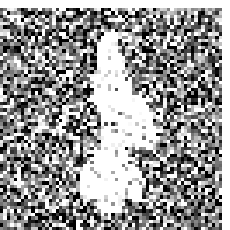}
 }
 \caption{The aircraft data set. Top row: Training set, middle row: test image set - 1 and bottom row: test image set - 2. Note that we remove the corresponding training image from the training set for each test image in our experiments. \label{fig:aircraft_training}}
\end{figure*}

\section{Experimental Results}
In this section, we present empirical results of our MCMC shape sampling algorithm on segmentation of potentially occluded objects in low-quality images. Note that, when dealing with segmentation of objects with unknown occlusions, $E_{data}(C)$ increases when the shape term delineates the boundaries in the occluded region. This can lead to overall increasing effect on $E(\mathcal{C}^{(t)})$ for a candidate curve and to the rejection of the candidate sample. In order to increase the acceptance rate of our approach, we use $\pi(C) \propto \exp(-E_{shape}(C))$ instead of $\pi(C) \propto \exp(-E(C))$ in our experiments involving occluded objects (see supplementary material for experiments involving missing data in which we use $\pi(C) \propto \exp(-E(C))$). This does not cause any problem in practice since the data fidelity term (together with the shape prior term) is involved in the curve perturbation step, enforcing consistency with the data.

We perform experiments on several data sets: aircraft~\cite{kim2007nonparametric}, MNIST handwritten digits~\cite{lecun1998gradient}, and walking silhouettes~\cite{cremers2006kernel}. In the following subsections, we present quantitative and visual results together with discussions of the experiments for each data set.

\subsection{Experiments on the aircraft data set}
The aircraft data set~\cite{kim2007nonparametric} contains 11 synthetically generated binary aircraft images as shown in the top row of Figure~\ref{fig:aircraft_training}. We construct the test images shown in the middle and the bottom rows of the same figure by cropping the left wings from the binary images to simulate occlusion and by adding different amounts of noise. Note that the test images shown in the middle row of Figure~\ref{fig:aircraft_training} (test image set - 1) have higher SNR than the ones shown in the bottom row (test image set - 2). In our experiments, we use this data set in leave-one-out fashion, i.e., we use one image as test and the remaining 10 binary images for training.
\begin{figure}[ht]
\centering
\subfigure[Test image set - 1]{
\begin{tabular}{cccc}
\multirow{3}{*}{\begin{tabular}[c]{@{}c@{}}Ground\\ Truth\end{tabular}} & \multirow{3}{*}{\begin{tabular}[c]{@{}c@{}}Result\\of~\cite{kim2007nonparametric}\end{tabular}} & \multirow{3}{*}{\begin{tabular}[c]{@{}c@{}}Best \\Sample\end{tabular}} & \multirow{3}{*}{\begin{tabular}[c]{@{}c@{}}PR Plots\end{tabular}} \\
                                                                        &                                                                                 &                                                                                                        &                                                                                     \\
                                                                        &                                                                                 &                                                                                                        &                                                                                     \\
\multicolumn{1}{@{}c@{}}{\includegraphics[width = 2cm, height = 2cm]{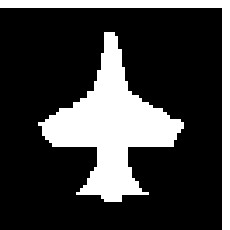}}                                                   &\multicolumn{1}{@{}c@{}}{\includegraphics[width = 2cm, height = 2cm]{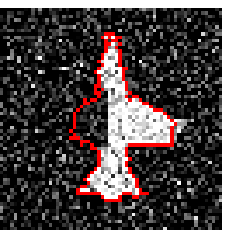}}                                                           & \multicolumn{1}{@{}c@{}}{\includegraphics[width = 2cm, height = 2cm]{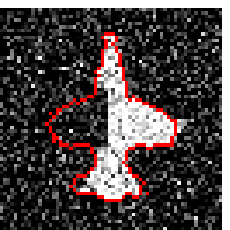}}                                                                                  & \multicolumn{1}{@{}c@{}}{\includegraphics[width = 2.5cm, height = 2cm]{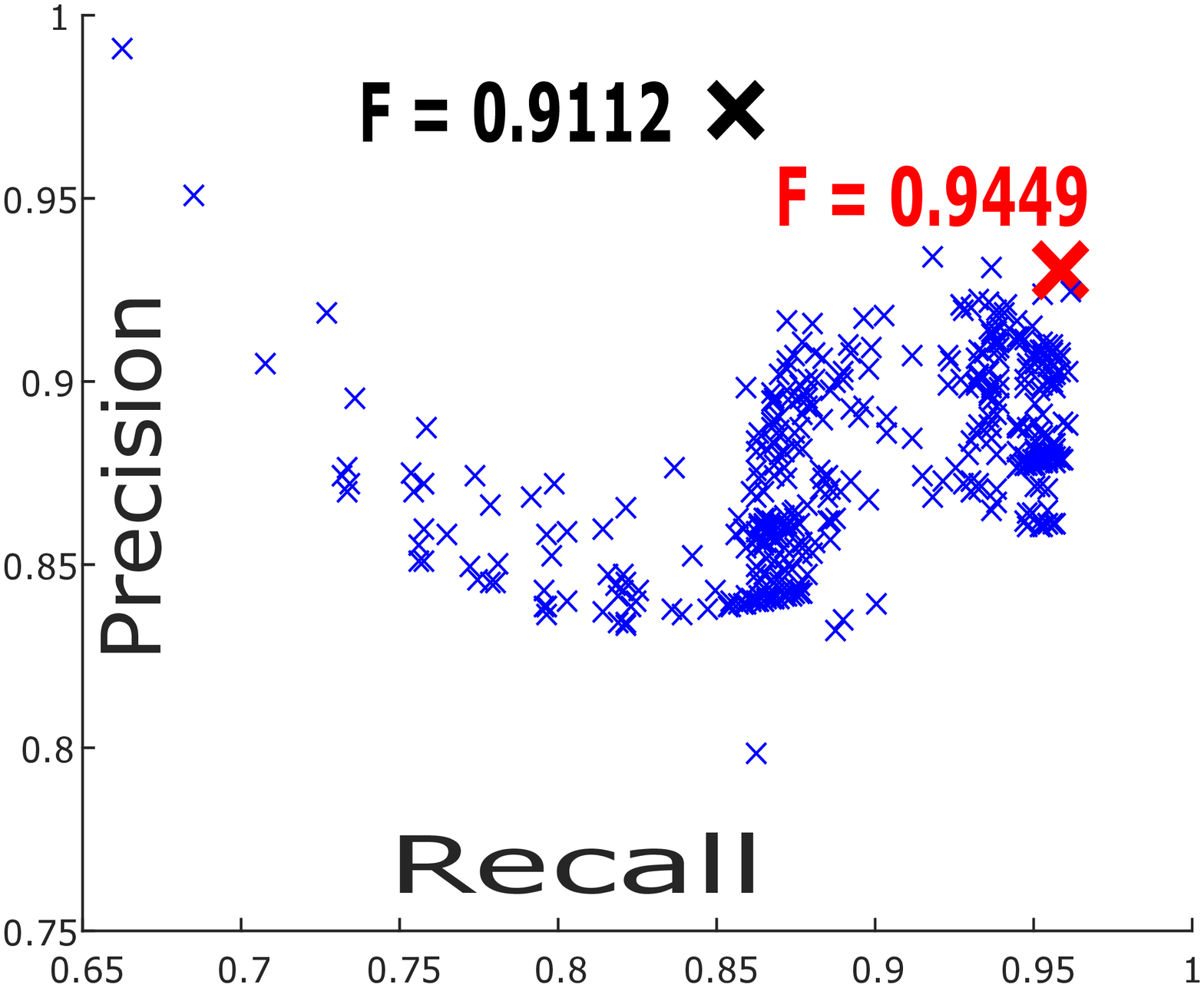}}

\\
\multicolumn{1}{@{}c@{}}{\includegraphics[width = 2cm, height = 2cm]{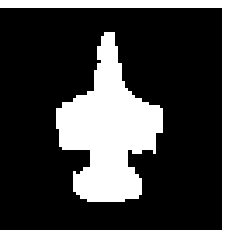}}                                                   &\multicolumn{1}{@{}c@{}}{\includegraphics[width = 2cm, height = 2cm]{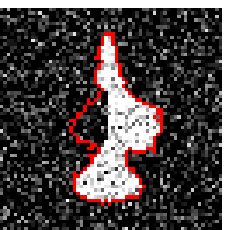}}                                                           & \multicolumn{1}{@{}c@{}}{\includegraphics[width = 2cm, height = 2cm]{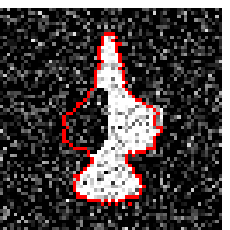}}                                                                                  & \multicolumn{1}{@{}c@{}}{\includegraphics[width = 2.5cm, height = 2cm]{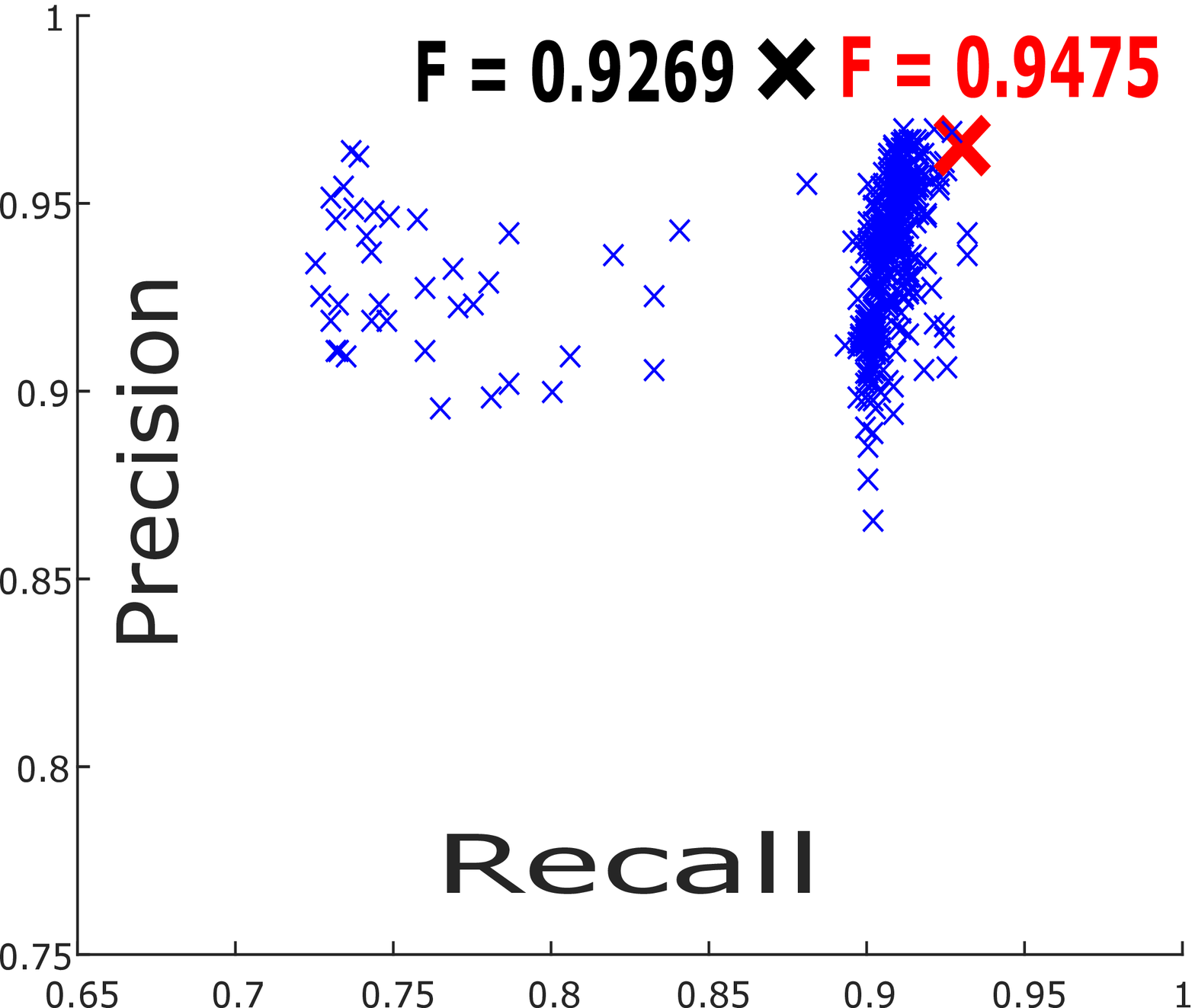}}

\\
\multicolumn{1}{@{}c@{}}{\includegraphics[width = 2cm, height = 2cm]{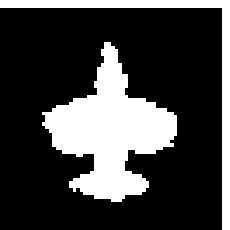}}                                                   &\multicolumn{1}{@{}c@{}}{\includegraphics[width = 2cm, height = 2cm]{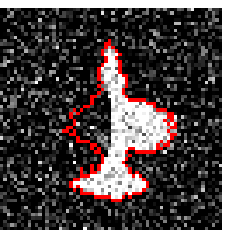}}                                                           & \multicolumn{1}{@{}c@{}}{\includegraphics[width = 2cm, height = 2cm]{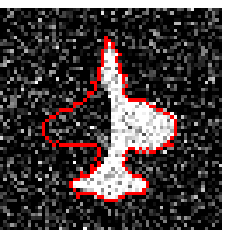}}                                                                                  & \multicolumn{1}{@{}c@{}}{\includegraphics[width = 2.5cm, height = 2cm]{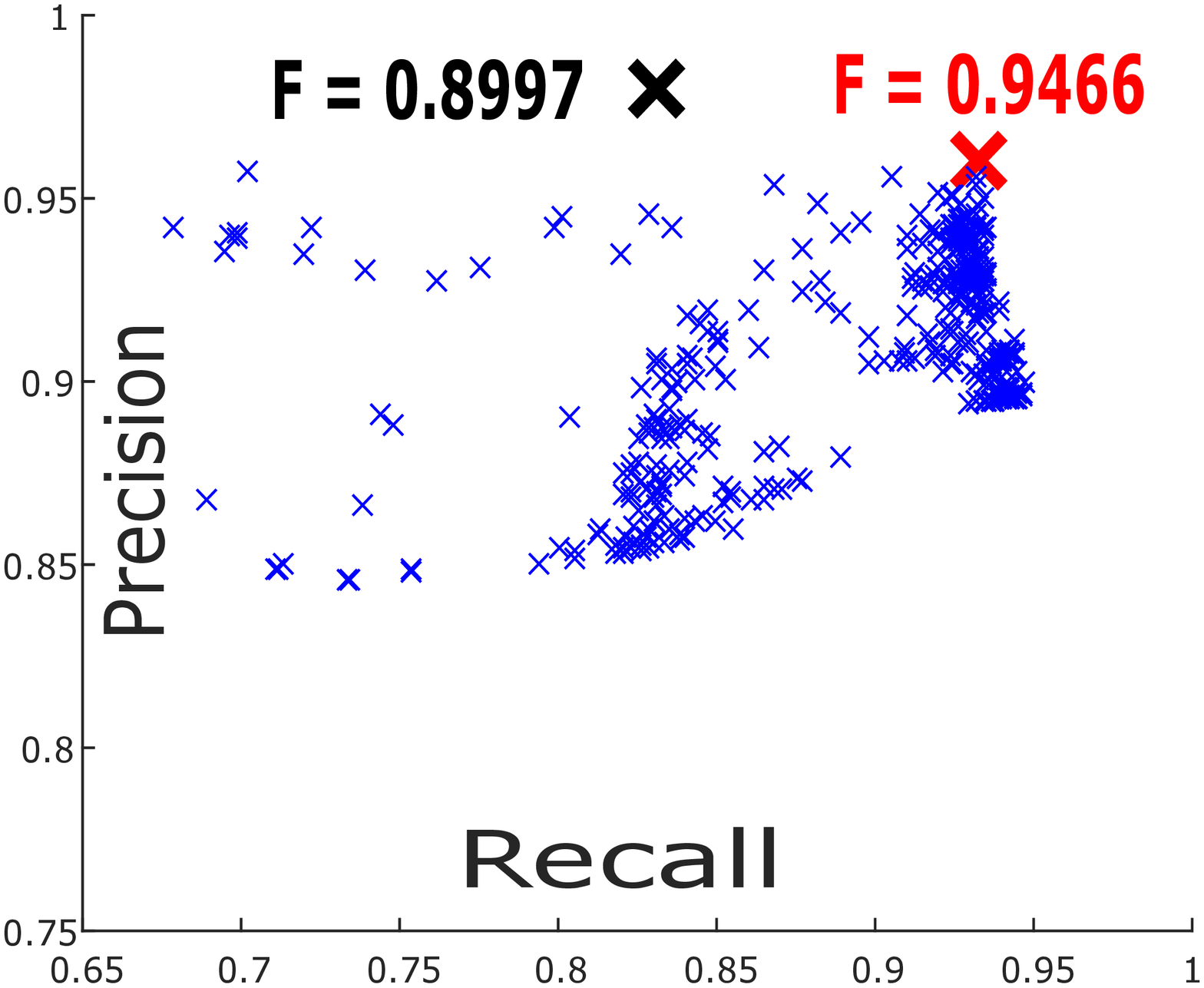}}                                                             
\end{tabular}
\label{fig:experiments_aircraft_test1}
 }

\subfigure[Test image set - 2]{
\begin{tabular}{cccc}
\multicolumn{1}{@{}c@{}}{\includegraphics[width = 2cm, height = 2cm]{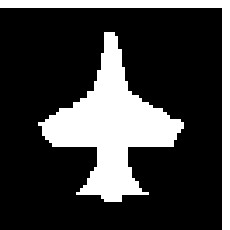}}                                                   &\multicolumn{1}{@{}c@{}}{\includegraphics[width = 2cm, height = 2cm]{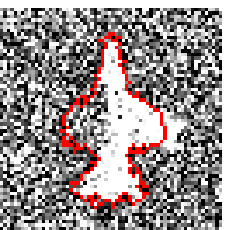}}                                                           & \multicolumn{1}{@{}c@{}}{\includegraphics[width = 2cm, height = 2cm]{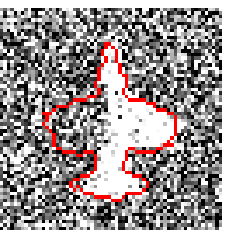}}                                                                                  & \multicolumn{1}{@{}c@{}}{\includegraphics[width = 2.5cm, height = 2cm]{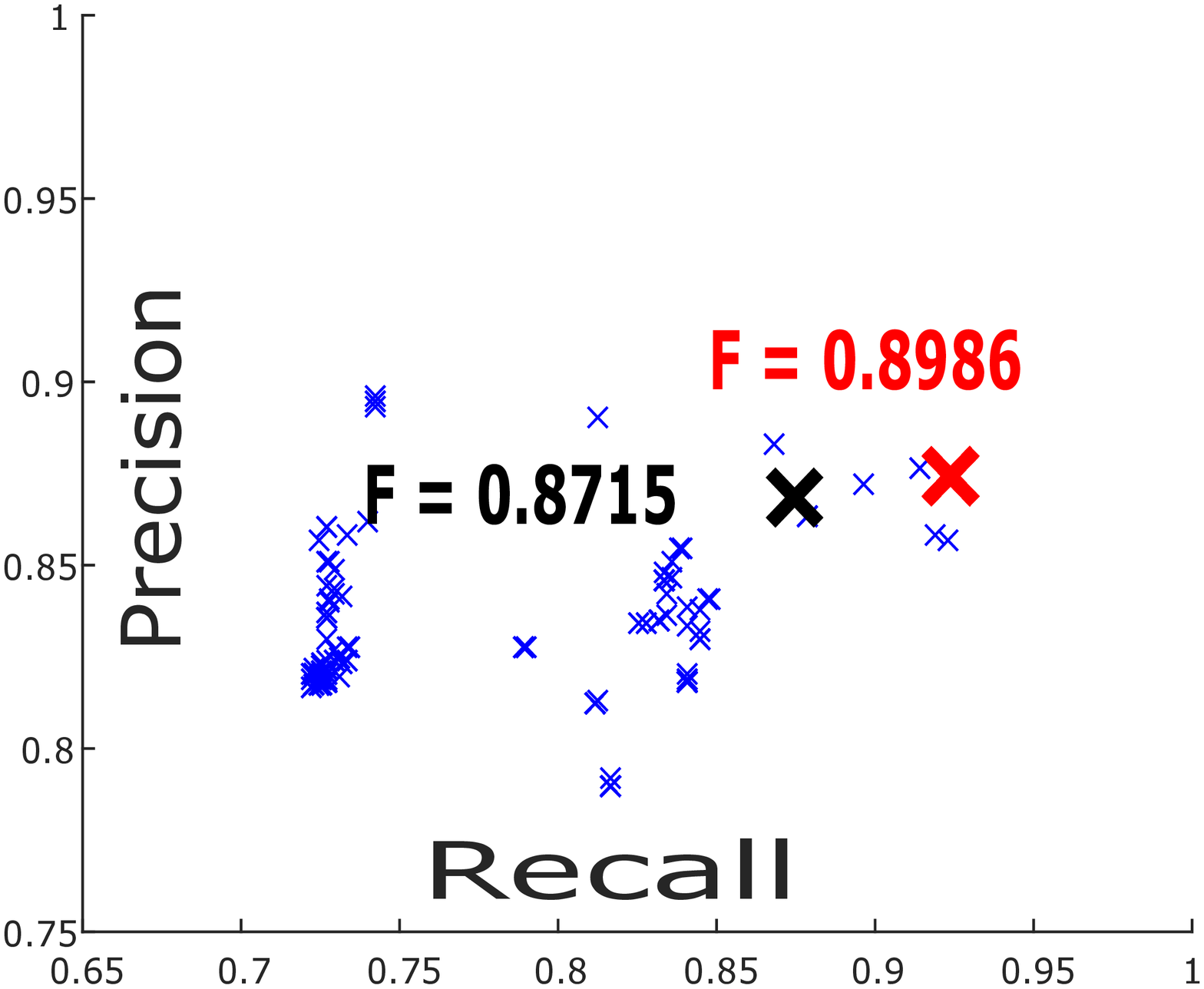}}

\\
\multicolumn{1}{@{}c@{}}{\includegraphics[width = 2cm, height = 2cm]{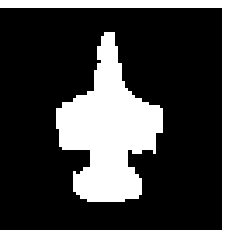}}                                                   &\multicolumn{1}{@{}c@{}}{\includegraphics[width = 2cm, height = 2cm]{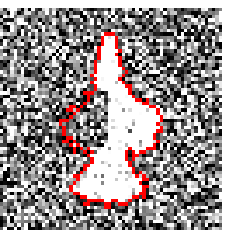}}                                                           & \multicolumn{1}{@{}c@{}}{\includegraphics[width = 2cm, height = 2cm]{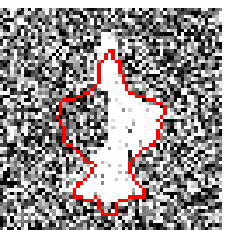}}                                                                                  & \multicolumn{1}{@{}c@{}}{\includegraphics[width = 2.5cm, height = 2cm]{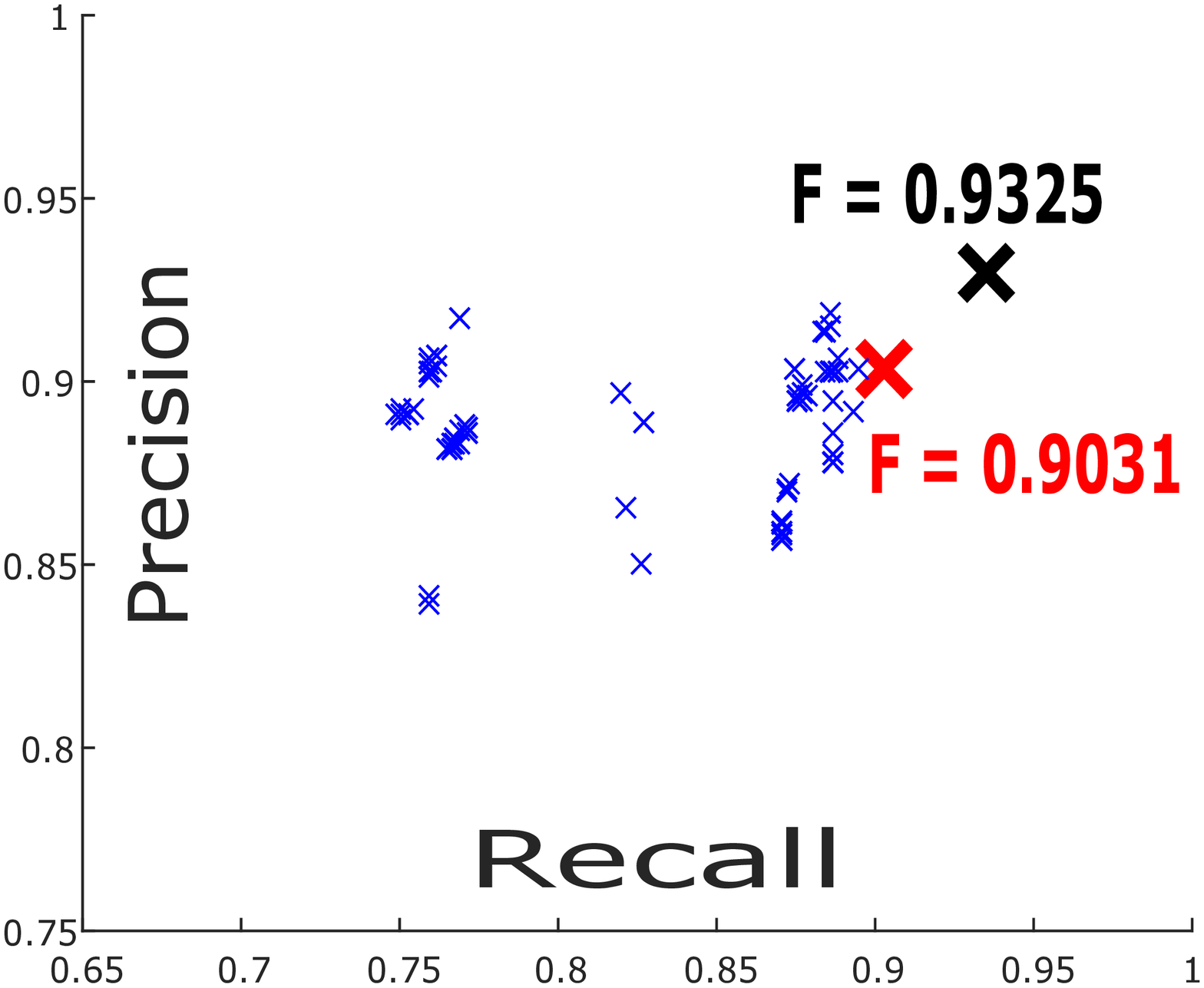}}

\\
\multicolumn{1}{@{}c@{}}{\includegraphics[width = 2cm, height = 2cm]{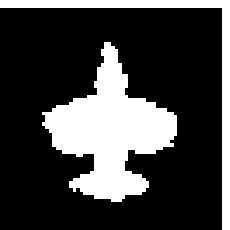}}                                                   &\multicolumn{1}{@{}c@{}}{\includegraphics[width = 2cm, height = 2cm]{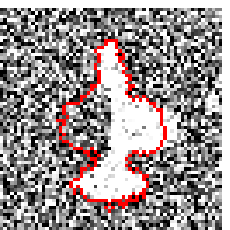}}                                                           & \multicolumn{1}{@{}c@{}}{\includegraphics[width = 2cm, height = 2cm]{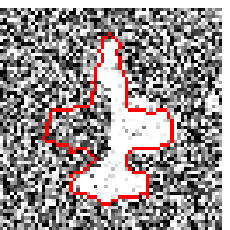}}                                                                                  & \multicolumn{1}{@{}c@{}}{\includegraphics[width = 2.5cm, height = 2cm]{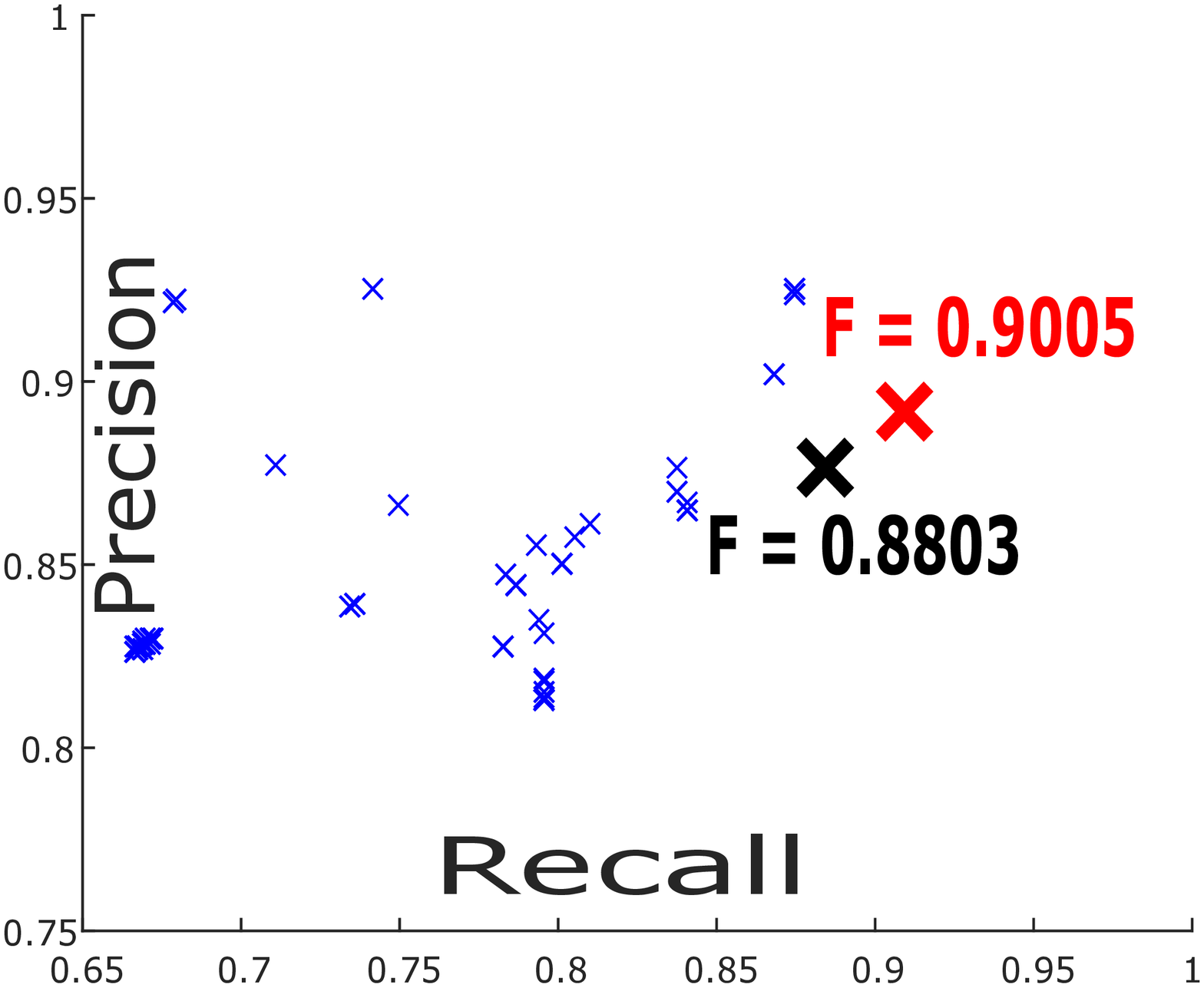}}                                                         
\end{tabular}
\label{fig:experiments_aircraft_test2}
 }
 \caption{Experiments on aircraft data set. Note that each row contains the results for a different test image. In the PR plots, \lq\textcolor{blue}{$\times$}\rq and \lq\textcolor{red}{$\times$}\rq mark the samples produced by our approach where \lq\textcolor{red}{$\times$}\rq indicates the sample with the best F-measure value, and \lq$\times$\rq marks that of segmentation of Kim et al.~\cite{kim2007nonparametric}.\label{fig:experiments_aircraft}}
\end{figure}

In Figure~\ref{fig:experiments_aircraft_test1}, we present some visual and quantitative results on the first three images from the test image set - 1 shown in Figure~\ref{fig:aircraft_training}. In this experiment, we generate 500 samples using our shape sampling approach for each test image. We also obtain segmentations using the optimization-based segmentation approach of Kim et al.~\cite{kim2007nonparametric} (see the second column of Figure~\ref{fig:experiments_aircraft_test1}). We compare each sample and the result of Kim et al.~\cite{kim2007nonparametric} with the corresponding ground truth image using precision - recall values and the F-measure. The samples with the best F-measure value are shown in the third column of Figure~\ref{fig:experiments_aircraft_test1}. Finally, we plot the precision - recall values (PR plots) for each sample and for the result of Kim et al.~\cite{kim2007nonparametric} in the fourth column of Figure~\ref{fig:experiments_aircraft_test1}. Here, the data fidelity term keeps the curve at the object boundaries and shape prior term helps to complete the shape in the occluded part. In our approach, since we select the most probable subset of training images and evolve the curve with the weighted average of these images, the results of our approach are more likely to produce better fits for the occluded part. In the experiments shown in Figure~\ref{fig:experiments_aircraft_test1}, our approach can generate better samples than the result of Kim et al.~\cite{kim2007nonparametric} in all test images. Moreover, our algorithm is able to generate many different samples in the solution space. By looking at these samples, one can also have more information about the confidence in a particular solution. 

We also perform experiments on the aircraft test image set - 2 shown in Figure~\ref{fig:aircraft_training} and present results on the first three images in Figure~\ref{fig:experiments_aircraft_test2}. The segmentation problem in this image set is more challenging than the previous case because of lower SNR. We perform experiments with the same settings as in test image set - 1 and present the results in the same way in Figure~\ref{fig:experiments_aircraft_test2}. In this case, we have to give more weight to the shape prior term during evolution to complete the occluded part because of the high amount of noise. Because of the limited role of the data fidelity term, the curve starts losing some part of the boundary after the shape term is turned on since the role of the data term is limited. Therefore, in this case, not only the occluded part but also the other parts of the aircraft shape approach a weighted average of the objects in the training set during curve evolution. Note from Figure~\ref{fig:experiments_aircraft_test2} that the results of Kim et al.~\cite{kim2007nonparametric} on different test images are very similar to one another. However, our sampling approach produces more diverse samples including better ones than the result of Kim et al.~\cite{kim2007nonparametric} in terms of F-measure in most cases. Additional results on all remaining test images shown in Figure~\ref{fig:aircraft_training} can be found in the supplementary material.

\subsection{Experiments on the MNIST data set}
\begin{figure}[h]
\centering
 \includegraphics[width = 1.2cm, height = 1.2cm]{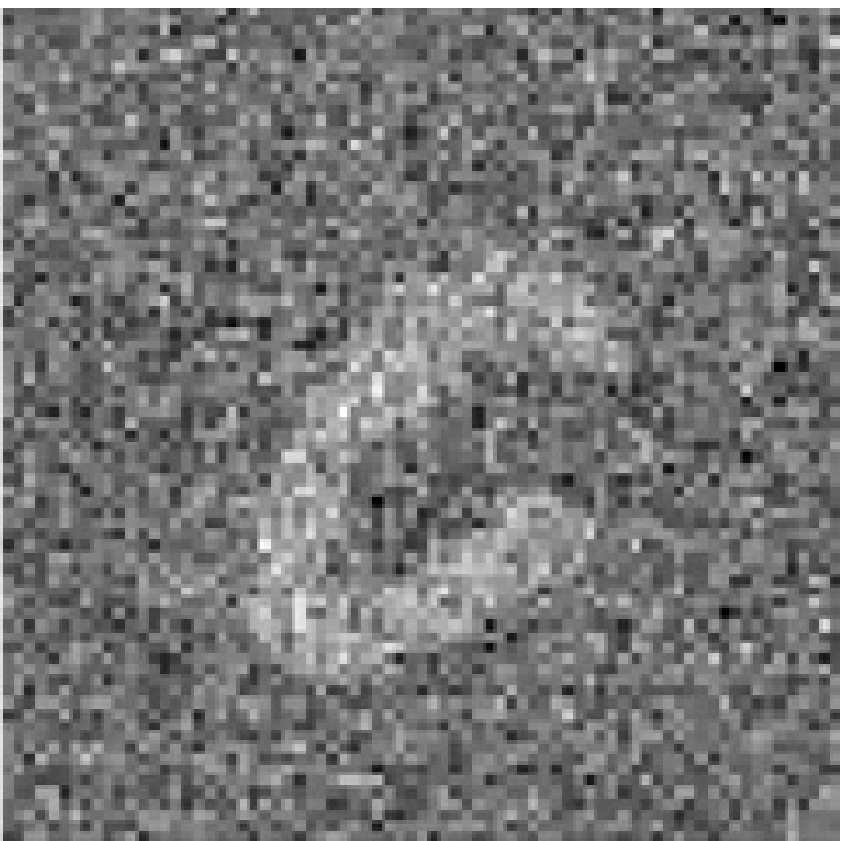}
 \quad
 \includegraphics[width = 1.2cm, height = 1.2cm]{figures/mnist_testimage2.eps}
 \quad
 \includegraphics[width = 1.2cm, height = 1.2cm]{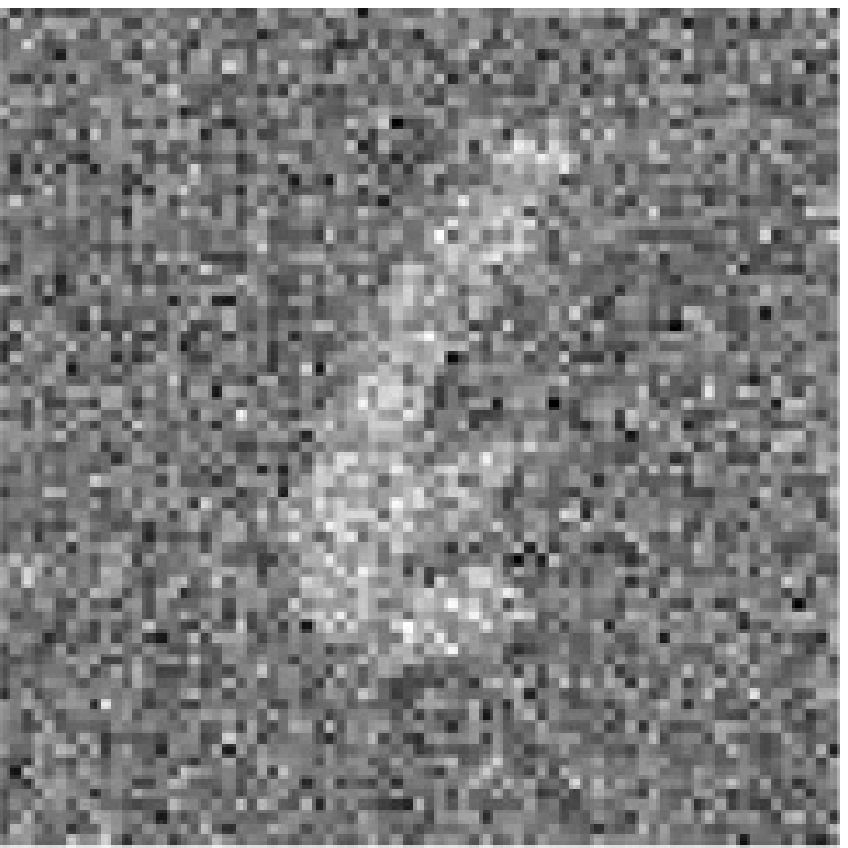}
 \caption{Test images from the MNIST data set. From left to right: MNIST - 1, MNIST - 2, and MNIST - 3. \label{fig:test_images}}
\end{figure}

In this section, we present empirical results on the MNIST handwritten digits~\cite{lecun1998gradient} data set which includes a multimodal shape density (i.e, training set contains shapes from multiple classes corresponding to different modes of the shape density). The MNIST handwritten digits data set contains 60,000 training examples and 10,000 test images from 10 different digit classes. In our experiments, we take a subset of 100 images for training such that each class contains 10 training examples. Test images, none of which are contained in the training set, are obtained by cropping some parts of the digits and adding noise. The test images that we use in our experiments are shown in Figure~\ref{fig:test_images}.

In our experiments on the MNIST data set, we generate 1000 samples using our shape sampling approach. In order to interpret our results, we use three methodologies: (1) Compute the average energy for each class by considering the samples generated in that class. Choose the best three classes with respect to average energy values. Display the best three samples from each class in terms of energy. These samples are most likely good representatives of the modes of the target distribution, (2) Compute the histogram images $H(x)$ which indicate in what percentage of the samples a particular pixel is inside the boundary. This can be simply computed by summing up all the binary samples and dividing by the number of samples~\cite{fan2007mcmc}. $H(x)$ can be computed for each class for problems involving multimodal shape densities. We draw the marginal confidence bounds, the bounds where $H(x) = 0.1$ and $H(x) = 0.9$, over the test image for each class, (3) Count the number of samples obtained from each class. This can allow a probabilistic interpretation of the results.
\begin{figure}[ht]
\centering
\begin{center}
\includegraphics[scale = 0.16]{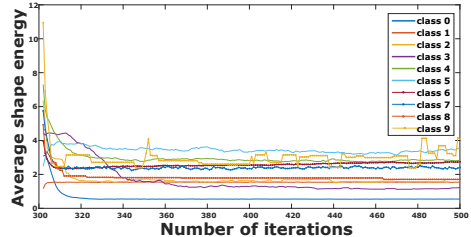}
\end{center}
\caption{Average shape energy ($E_{shape}(C)$) across all sampling iterations for all digit classes for test image MNIST - 1. Note that the number of iterations start from 300 in $x$-axis because the previous iterations involve segmentation with the data term only.\label{fig:average_energy_mnist1}}
\end{figure}

Figure~\ref{fig:average_energy_mnist1} demonstrates the average shape energy for each class, $E_{shape}(C)$, as a function of sampling iterations for test image MNIST - 1. We note that while the average energy appears to be smoothly converging, the energy for each sample path can sharply increase and decrease. The plot of class 9 in Figure~\ref{fig:average_energy_mnist1} exhibits such a such pattern because there is only one sample generated from this class. As the number of samples generated in each class increases, the average sample path converges to a stationary distribution.

\begin{table}[ht]
\centering
\resizebox{.40\textwidth}{!}{
\begin{tabular}{|c|c|c|c|c|c|c|c|c|c|c|}
\hline
\multirow{2}{*}{\textbf{Test Image}}&\multicolumn{10}{|c|}{\textbf{Digit Class}} \\ \cline{2-11}
	&\textbf{0}	&\textbf{1}	&\textbf{2}	&\textbf{3}	&\textbf{4}	&\textbf{5}	&\textbf{6}	&\textbf{7}	&\textbf{8}	&\textbf{9} \\ \hline
\textbf{MNIST - 1}			&336	&433	&6	&18	&29	&38	&115	&16	&8	&1 \\ \hline
\textbf{MNIST - 2}			&4	&691	&8	&3	&96	&9	&0	&120	&3	&66 \\ \hline
\textbf{MNIST - 3}			&119&661	&8	&1	&2	&11	&154	&14	&28	&2 \\ \hline
\end{tabular}
}
\caption{Number of samples generated for each digit class in test images from the MNIST data set.\label{tab:numberOfSamples}}
\end{table}
Number of samples generated from each digit class for all the three test images is shown in Table~\ref{tab:numberOfSamples}. This allows us to make a probabilistic interpretation of the segmentation results. One can evaluate the confidence level of the results by analyzing the number of samples generated from a class over all samples. 

\begin{figure*}[ht]
\centering
\begin{tabular}{cccccccccccc}
\begin{tabular}[c]{@{}c@{}}MCB\\ image\end{tabular} & \multicolumn{3}{c}{The best 3 samples} & \begin{tabular}[c]{@{}c@{}}MCB\\ image\end{tabular} & \multicolumn{3}{c}{The best 3 samples} & \begin{tabular}[c]{@{}c@{}}MCB \\ image\end{tabular} & \multicolumn{3}{c}{The best 3 samples}
\\
\includegraphics[width = 1.2cm, height = 1.2cm]{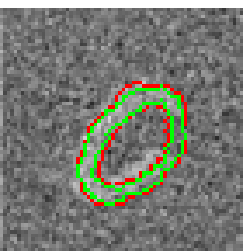}&
\multicolumn{1}{@{}c@{}}{\includegraphics[width = 1.2cm, height = 1.2cm]{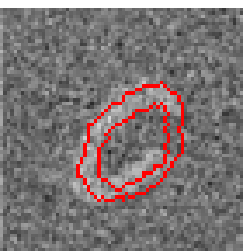}}&
\multicolumn{1}{@{}c@{}}{\includegraphics[width = 1.2cm, height = 1.2cm]{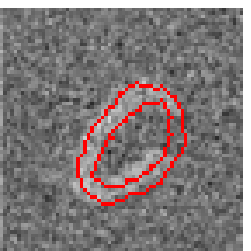}}& 
\multicolumn{1}{@{}c@{}}{\includegraphics[width = 1.2cm, height = 1.2cm]{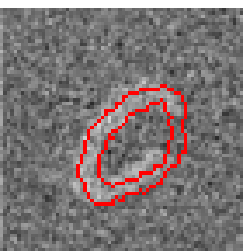}}&
\quad
\includegraphics[width = 1.2cm, height = 1.2cm]{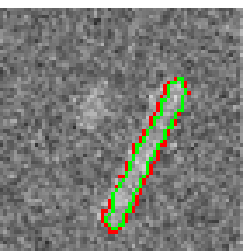}&
\multicolumn{1}{@{}c@{}}{\includegraphics[width = 1.2cm, height = 1.2cm]{figures/mnist2_1_1.eps}}&
\multicolumn{1}{@{}c@{}}{\includegraphics[width = 1.2cm, height = 1.2cm]{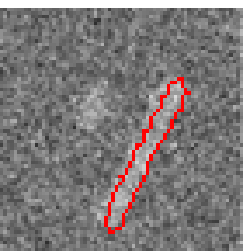}}&
\multicolumn{1}{@{}c@{}}{\includegraphics[width = 1.2cm, height = 1.2cm]{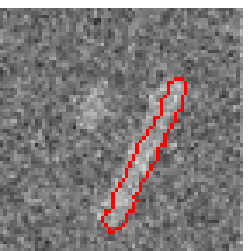}}&
\quad
\includegraphics[width = 1.2cm, height = 1.2cm]{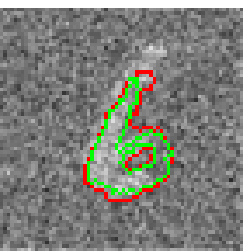}&
\multicolumn{1}{@{}c@{}}{\includegraphics[width = 1.2cm, height = 1.2cm]{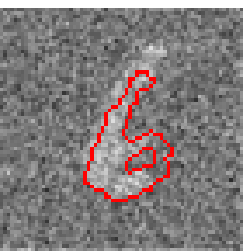}}&
\multicolumn{1}{@{}c@{}}{\includegraphics[width = 1.2cm, height = 1.2cm]{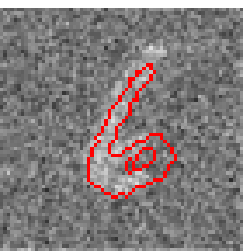}}&
\multicolumn{1}{@{}c@{}}{\includegraphics[width = 1.2cm, height = 1.2cm]{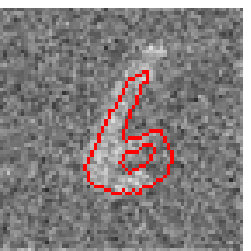}}
\\
\includegraphics[width = 1.2cm, height = 1.2cm]{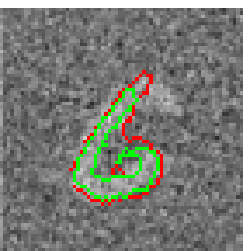}&
\multicolumn{1}{@{}c@{}}{\includegraphics[width = 1.2cm, height = 1.2cm]{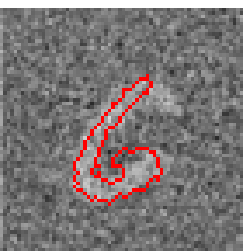}}&
\multicolumn{1}{@{}c@{}}{\includegraphics[width = 1.2cm, height = 1.2cm]{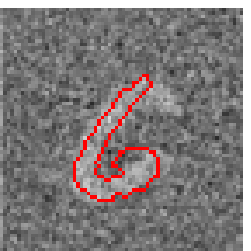}}&
\multicolumn{1}{@{}c@{}}{\includegraphics[width = 1.2cm, height = 1.2cm]{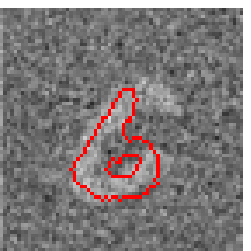}}&
\quad
\includegraphics[width = 1.2cm, height = 1.2cm]{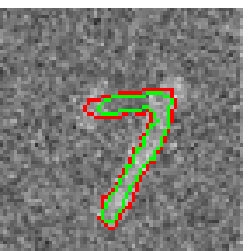}&
\multicolumn{1}{@{}c@{}}{\includegraphics[width = 1.2cm, height = 1.2cm]{figures/mnist2_7_1.eps}}&
\multicolumn{1}{@{}c@{}}{\includegraphics[width = 1.2cm, height = 1.2cm]{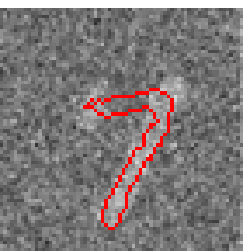}}&
\multicolumn{1}{@{}c@{}}{\includegraphics[width = 1.2cm, height = 1.2cm]{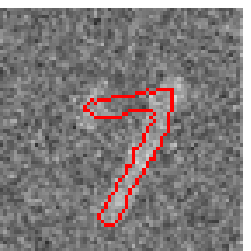}}&
\quad
\includegraphics[width = 1.2cm, height = 1.2cm]{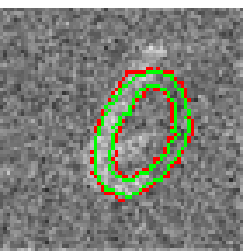}&
\multicolumn{1}{@{}c@{}}{\includegraphics[width = 1.2cm, height = 1.2cm]{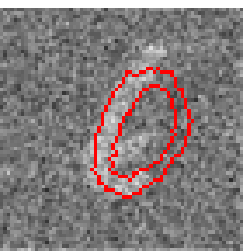}}&
\multicolumn{1}{@{}c@{}}{\includegraphics[width = 1.2cm, height = 1.2cm]{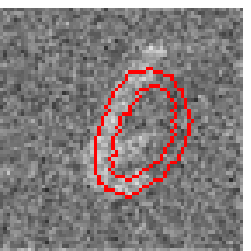}}& 
\multicolumn{1}{@{}c@{}}{\includegraphics[width = 1.2cm, height = 1.2cm]{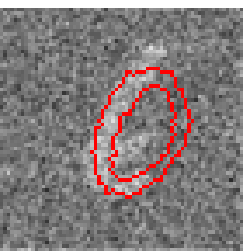}}
\\
\includegraphics[width = 1.2cm, height = 1.2cm]{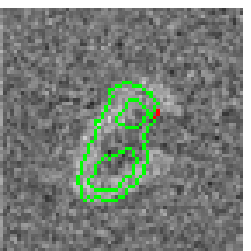}&
\multicolumn{1}{@{}c@{}}{\includegraphics[width = 1.2cm, height = 1.2cm]{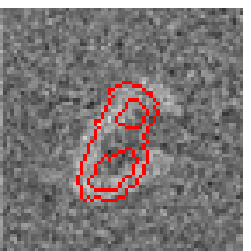}}&
\multicolumn{1}{@{}c@{}}{\includegraphics[width = 1.2cm, height = 1.2cm]{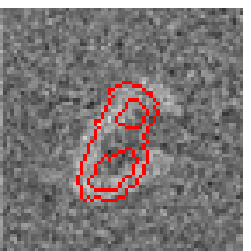}}&
\multicolumn{1}{@{}c@{}}{\includegraphics[width = 1.2cm, height = 1.2cm]{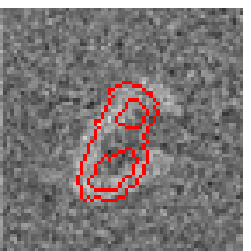}}&
\quad
\includegraphics[width = 1.2cm, height = 1.2cm]{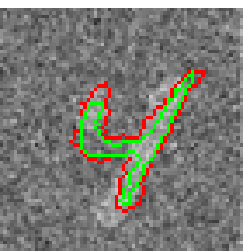}&
\multicolumn{1}{@{}c@{}}{\includegraphics[width = 1.2cm, height = 1.2cm]{figures/mnist2_4_1.eps}}&
\multicolumn{1}{@{}c@{}}{\includegraphics[width = 1.2cm, height = 1.2cm]{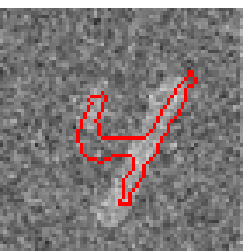}}& 
\multicolumn{1}{@{}c@{}}{\includegraphics[width = 1.2cm, height = 1.2cm]{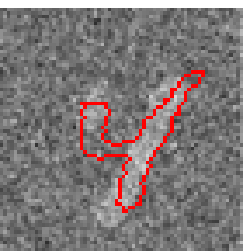}}&
\quad
\includegraphics[width = 1.2cm, height = 1.2cm]{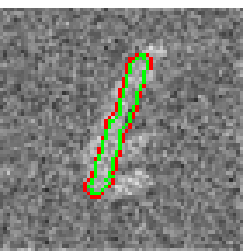}&
\multicolumn{1}{@{}c@{}}{\includegraphics[width = 1.2cm, height = 1.2cm]{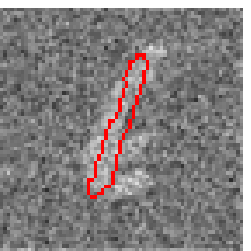}}&
\multicolumn{1}{@{}c@{}}{\includegraphics[width = 1.2cm, height = 1.2cm]{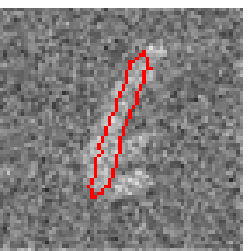}}& 
\multicolumn{1}{@{}c@{}}{\includegraphics[width = 1.2cm, height = 1.2cm]{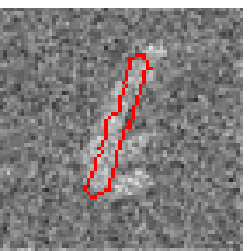}}
\\
\multicolumn{4}{c}{MNIST - 1}                                                                & \multicolumn{4}{c}{MNIST - 2}                                                                & \multicolumn{4}{c}{MNIST - 3}                                                             
\label{fig:results_MNIST}
\end{tabular}
 \caption{Experiments on the MNIST data set. Note that in MCB images, red and green contours are the marginal confidence bounds at $H(x) = 0.1$ and $H(x) = 0.9$, respectively.
 \label{fig:result_MNIST}}
\end{figure*}

In different segmentation applications, one can investigate solutions obtained from different parts of the posterior probability density. Especially, in the case of multimodal shape densities, segmentation results obtained from multiple modes might be interesting and might offer reasonable solutions. Figure~\ref{fig:result_MNIST} shows some visual results obtained from the experiments on the MNIST data set. For each test image, we display the results from the best three digit classes where, the quality of each class is computed as the average energy, $E(C)$, of the samples in that class. Also, for each class, we show three samples having the best energy values. These results show that our algorithm is able to find reasonable solutions from different modes of the posterior density. In Figure~\ref{fig:result_MNIST}, we also present marginal confidence bounds (MCB images) obtained from the samples in each class. The figure demonstrates the marginal confidence bounds at different levels of the histogram image, $H(x)$, for the best classes in all test images. $H(x) = 0.1$ and $H(x) = 0.9$ indicate the low probability and the high probability regions, respectively.

\subsection{Experiments on the walking silhouettes data set}
\begin{figure}[ht]
\centering
\subfigure{
 \includegraphics[width = 0.8cm, height = 1.4cm]{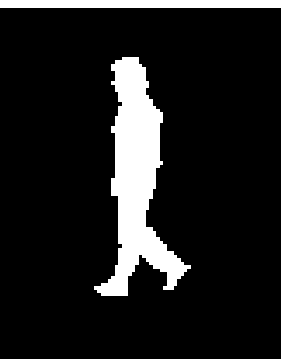}
 \includegraphics[width = 0.8cm, height = 1.4cm]{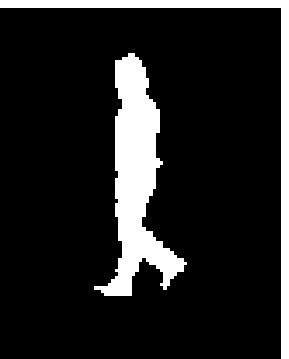}
 \includegraphics[width = 0.8cm, height = 1.4cm]{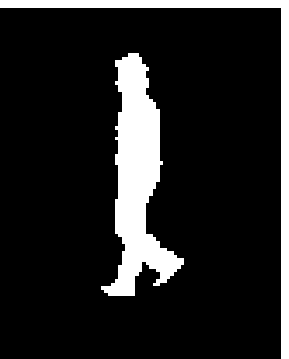}
 \includegraphics[width = 0.8cm, height = 1.4cm]{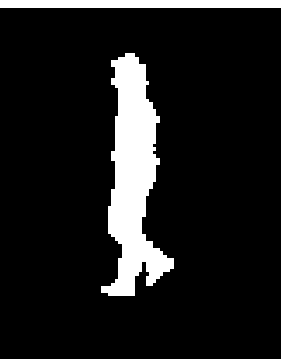}
 \includegraphics[width = 0.8cm, height = 1.4cm]{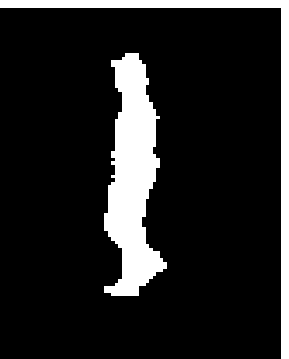}
 \includegraphics[width = 0.8cm, height = 1.4cm]{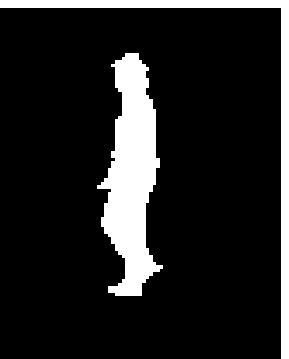}
 \includegraphics[width = 0.8cm, height = 1.4cm]{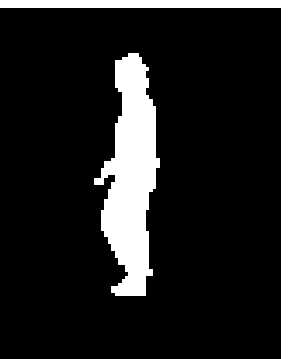}
 \includegraphics[width = 0.8cm, height = 1.4cm]{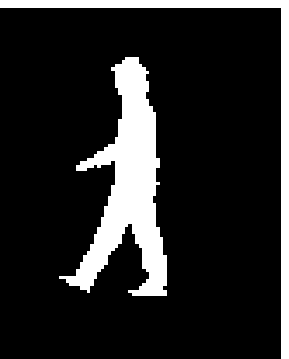}
 }
 \subfigure{
 \includegraphics[width = 0.8cm, height = 1.4cm]{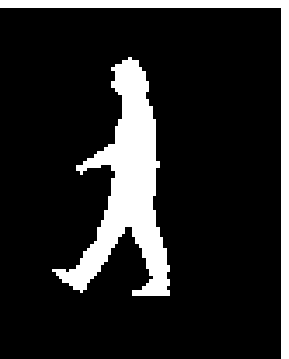}
 \includegraphics[width = 0.8cm, height = 1.4cm]{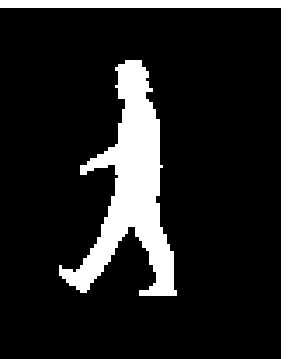}
 \includegraphics[width = 0.8cm, height = 1.4cm]{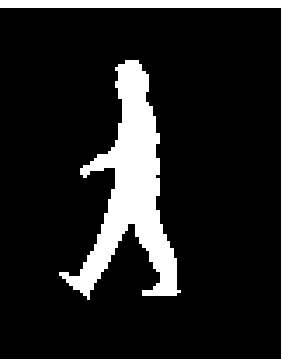}
 \includegraphics[width = 0.8cm, height = 1.4cm]{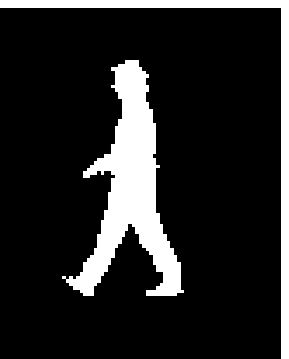}
 \includegraphics[width = 0.8cm, height = 1.4cm]{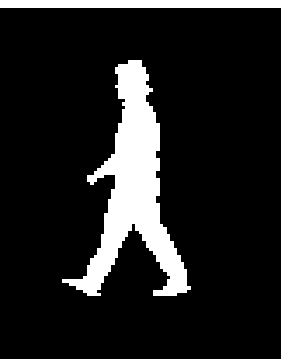}
 \includegraphics[width = 0.8cm, height = 1.4cm]{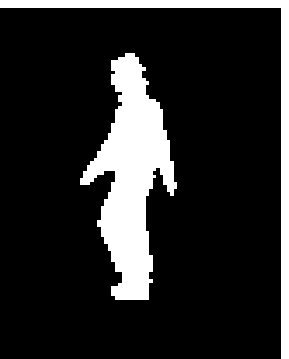}
 \includegraphics[width = 0.8cm, height = 1.4cm]{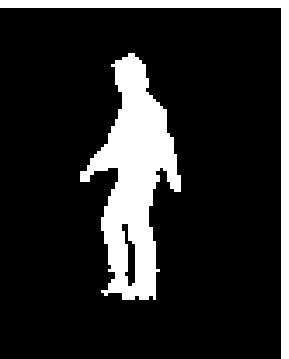}
 \includegraphics[width = 0.8cm, height = 1.4cm]{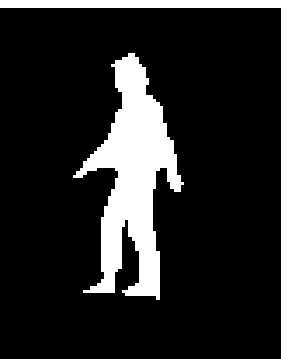}
 }
 \caption{The training set for the walking silhouettes data set. \label{fig:training_walking}}
\end{figure}

In this experiment, we test the performance of local shape priors extension of our MCMC shape sampling approach and compare it with the one that uses global shape priors, as well as with the method of Kim et al.~\cite{kim2007nonparametric}. We choose a subset of 30 binary images of a walking person from the walking silhouettes data set~\cite{cremers2006kernel}. A subset of 16 images shown in Figure~\ref{fig:training_walking} among these 30 binary images are used for training. The remaining 14 binary images are used to construct test images by adding a high amount of noise. 

For the sake of brevity, we present results on 3 test images in Figure~\ref{fig:experiments_walking}. Additional results can be found in the supplementary material. Similar to the evaluations performed for the aircraft data set, we plot the PR values for each sample obtained by our approaches (with global and local priors) and by the approach of Kim et al.~\cite{kim2007nonparametric}. According to the results, our proposed approach with global shape priors produces samples that have F-measure values better than or equal to the result of Kim et al.~\cite{kim2007nonparametric} in all test images. By using local shape priors, we can generate even better samples than both Kim et al.~\cite{kim2007nonparametric} and the approach with global shape priors. Moreover, it seems that our approach based on local shape priors is able to sample the space more effectively than the approach with global shape priors.
\begin{figure*}[ht]
\centering
\begin{tabular}{ccccccc}
\begin{tabular}[c]{@{}c@{}}Test\\ Image\end{tabular} & \begin{tabular}[c]{@{}c@{}}Ground\\ Truth\end{tabular} & \begin{tabular}[c]{@{}c@{}}Result\\ of~\cite{kim2007nonparametric}\end{tabular} & \begin{tabular}[c]{@{}c@{}}Best sample\\obtained using\\ global priors\end{tabular} & \begin{tabular}[c]{@{}c@{}}PR plots with\\ global priors\end{tabular} & \begin{tabular}[c]{@{}c@{}}Best sample\\ obtained using\\ local priors\end{tabular} & \begin{tabular}[c]{@{}c@{}}PR plots with\\ local priors\end{tabular} \\

\multicolumn{1}{@{}c@{}}{\includegraphics[width = 1.2cm, height = 2cm]{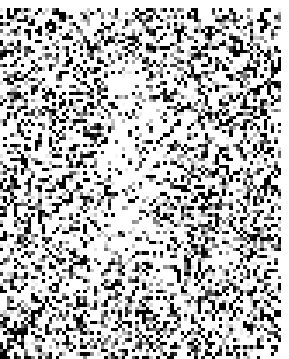}                                             } & \multicolumn{1}{@{}c@{}}{\includegraphics[width = 1.2cm, height = 2cm]{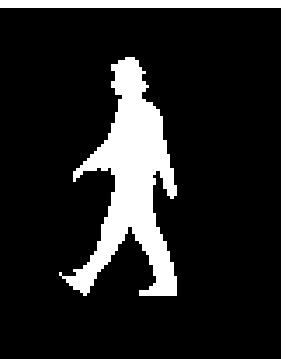}                                                     } & \multicolumn{1}{@{}c@{}}{\includegraphics[width = 1.2cm, height = 2cm]{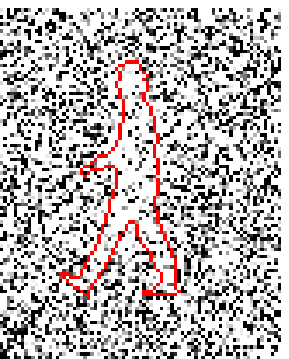}                                                                                } & \multicolumn{1}{@{}c@{}}{\includegraphics[width = 1.2cm, height = 2cm]{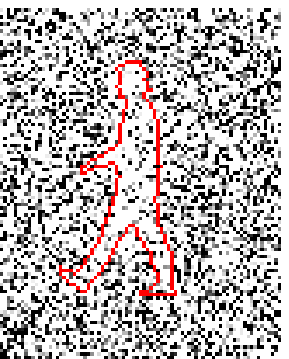}                                                                                 } & \multicolumn{1}{@{}c@{}}{\includegraphics[width = 3.2cm, height = 2cm]{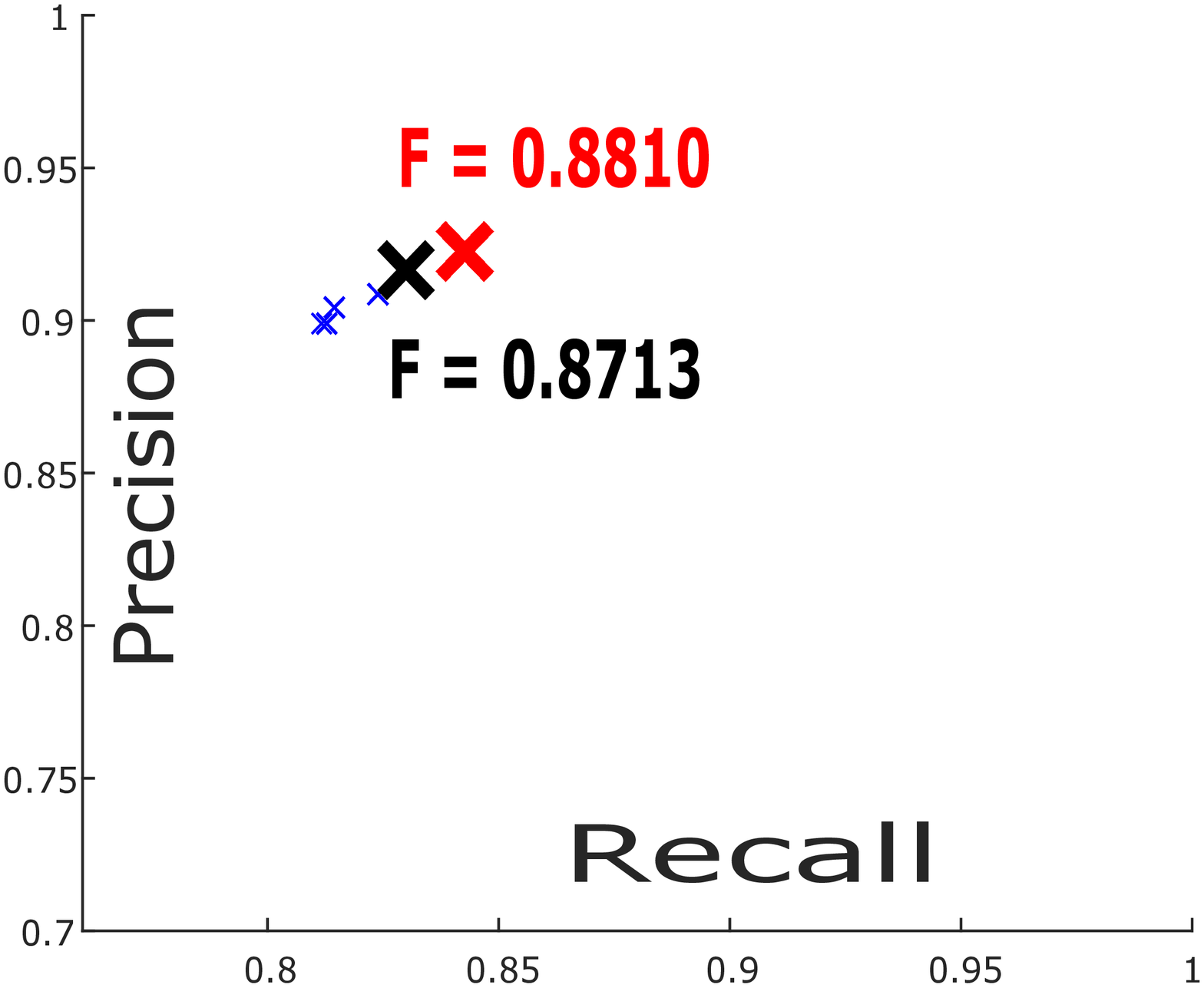}}                                                                     & \multicolumn{1}{@{}c@{}}{\includegraphics[width = 1.2cm, height = 2cm]{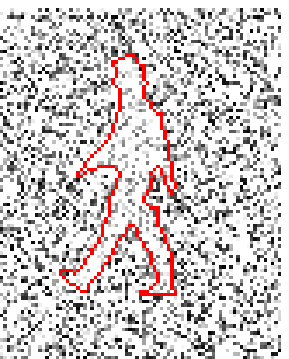}                                                                                } & \multicolumn{1}{@{}c@{}}{\includegraphics[width = 3.2cm, height = 2cm]{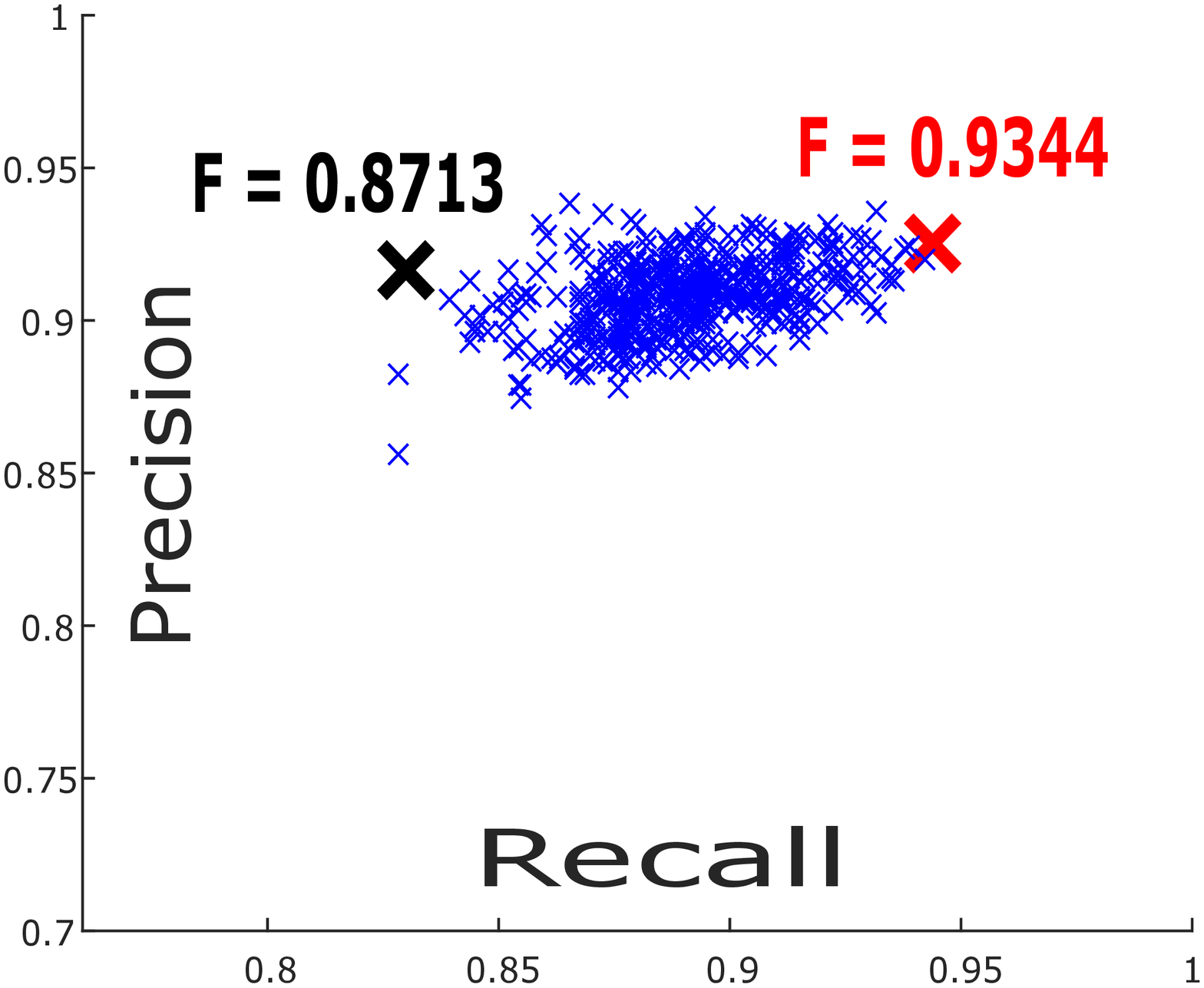}}
\\
\multicolumn{1}{@{}c@{}}{\includegraphics[width = 1.2cm, height = 2cm]{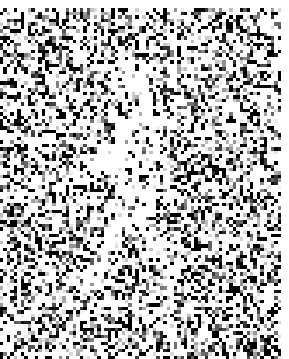}                                             } & \multicolumn{1}{@{}c@{}}{\includegraphics[width = 1.2cm, height = 2cm]{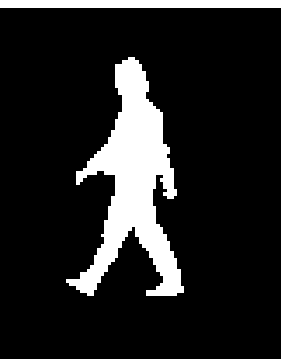}                                                     } & \multicolumn{1}{@{}c@{}}{\includegraphics[width = 1.2cm, height = 2cm]{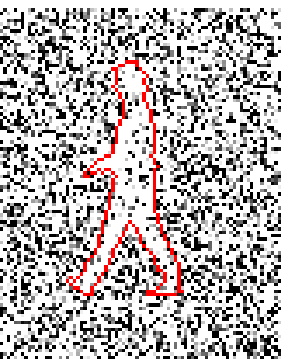}                                                                                } & \multicolumn{1}{@{}c@{}}{\includegraphics[width = 1.2cm, height = 2cm]{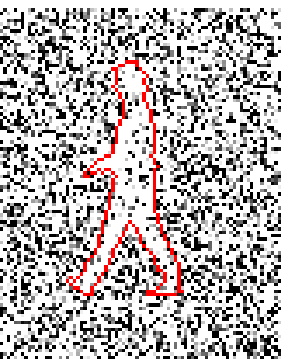}                                                                                 } & \multicolumn{1}{@{}c@{}}{\includegraphics[width = 3.2cm, height = 2cm]{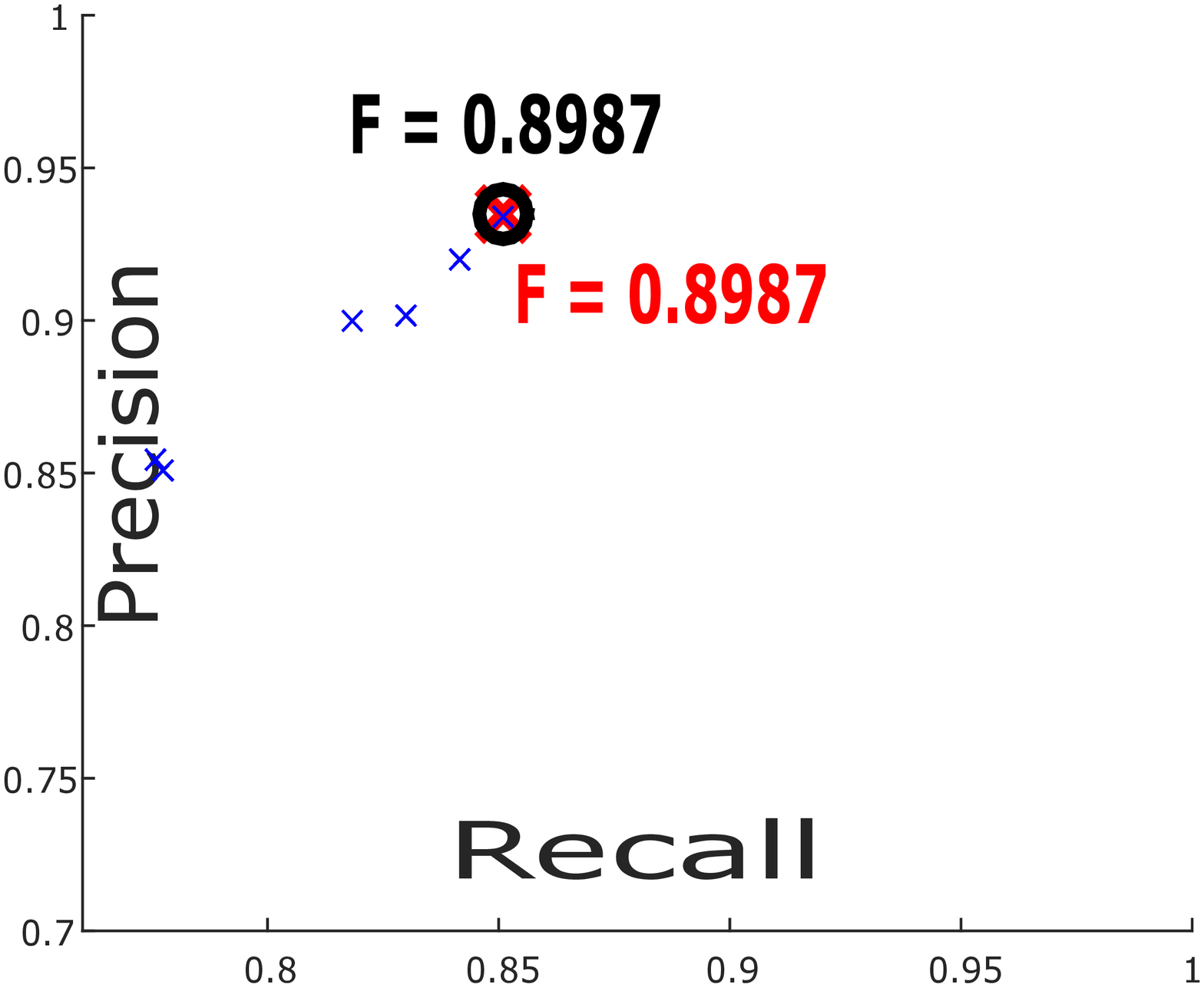}}                                                                     & \multicolumn{1}{@{}c@{}}{\includegraphics[width = 1.2cm, height = 2cm]{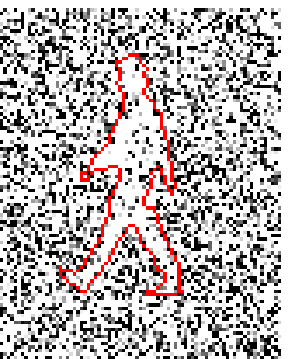}                                                                                } & \multicolumn{1}{@{}c@{}}{\includegraphics[width = 3.2cm, height = 2cm]{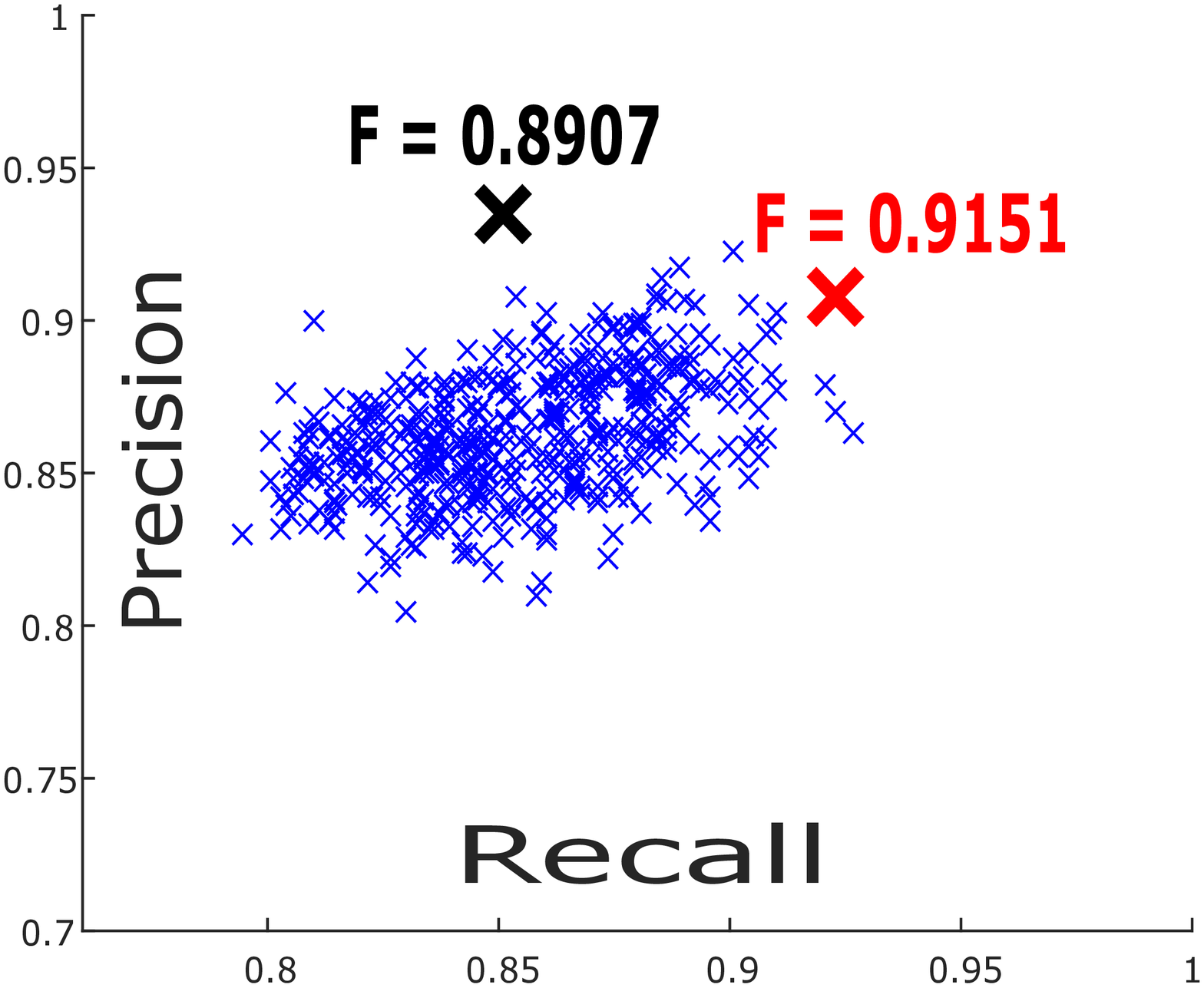}                                                                   }
\\
\multicolumn{1}{@{}c@{}}{\includegraphics[width = 1.2cm, height = 2cm]{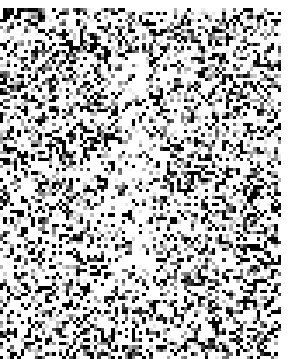}                                             } & \multicolumn{1}{@{}c@{}}{\includegraphics[width = 1.2cm, height = 2cm]{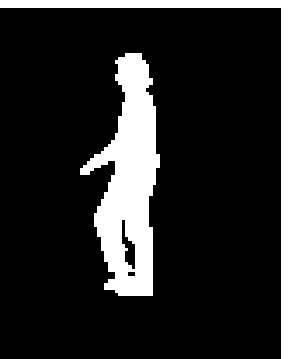}                                                     } & \multicolumn{1}{@{}c@{}}{\includegraphics[width = 1.2cm, height = 2cm]{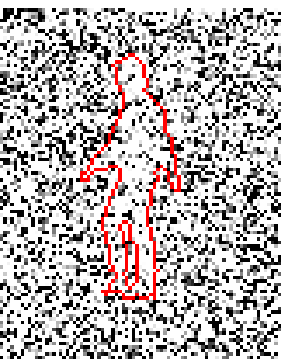}                                                                                } & \multicolumn{1}{@{}c@{}}{\includegraphics[width = 1.2cm, height = 2cm]{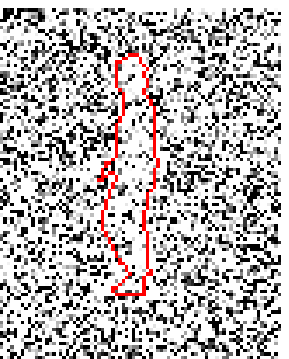}                                                                                 } & \multicolumn{1}{@{}c@{}}{\includegraphics[width = 3.2cm, height = 2cm]{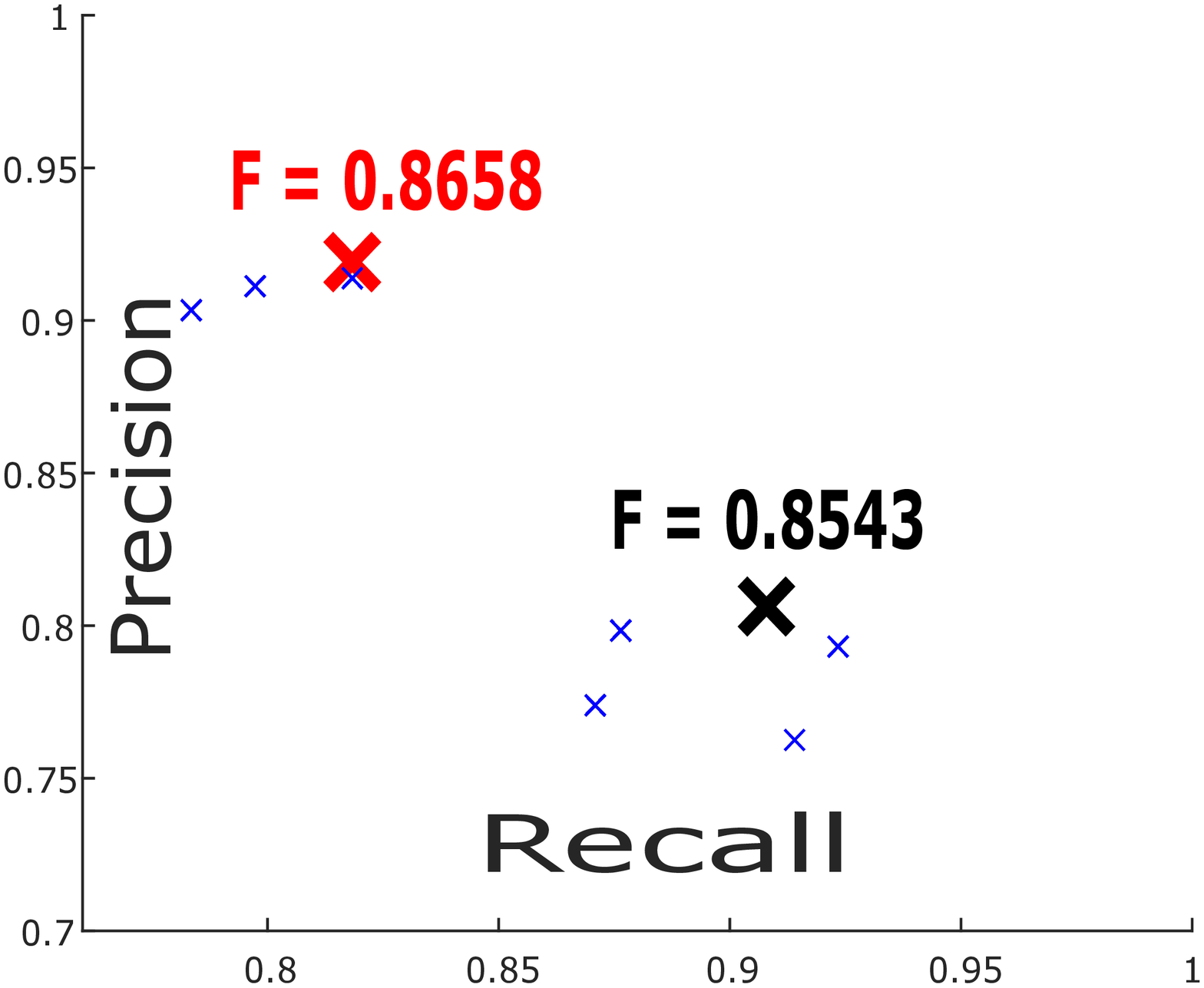}}                                                                     & \multicolumn{1}{@{}c@{}}{\includegraphics[width = 1.2cm, height = 2cm]{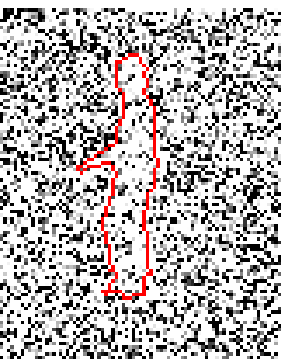}                                                                                } & \multicolumn{1}{@{}c@{}}{\includegraphics[width = 3.2cm, height = 2cm]{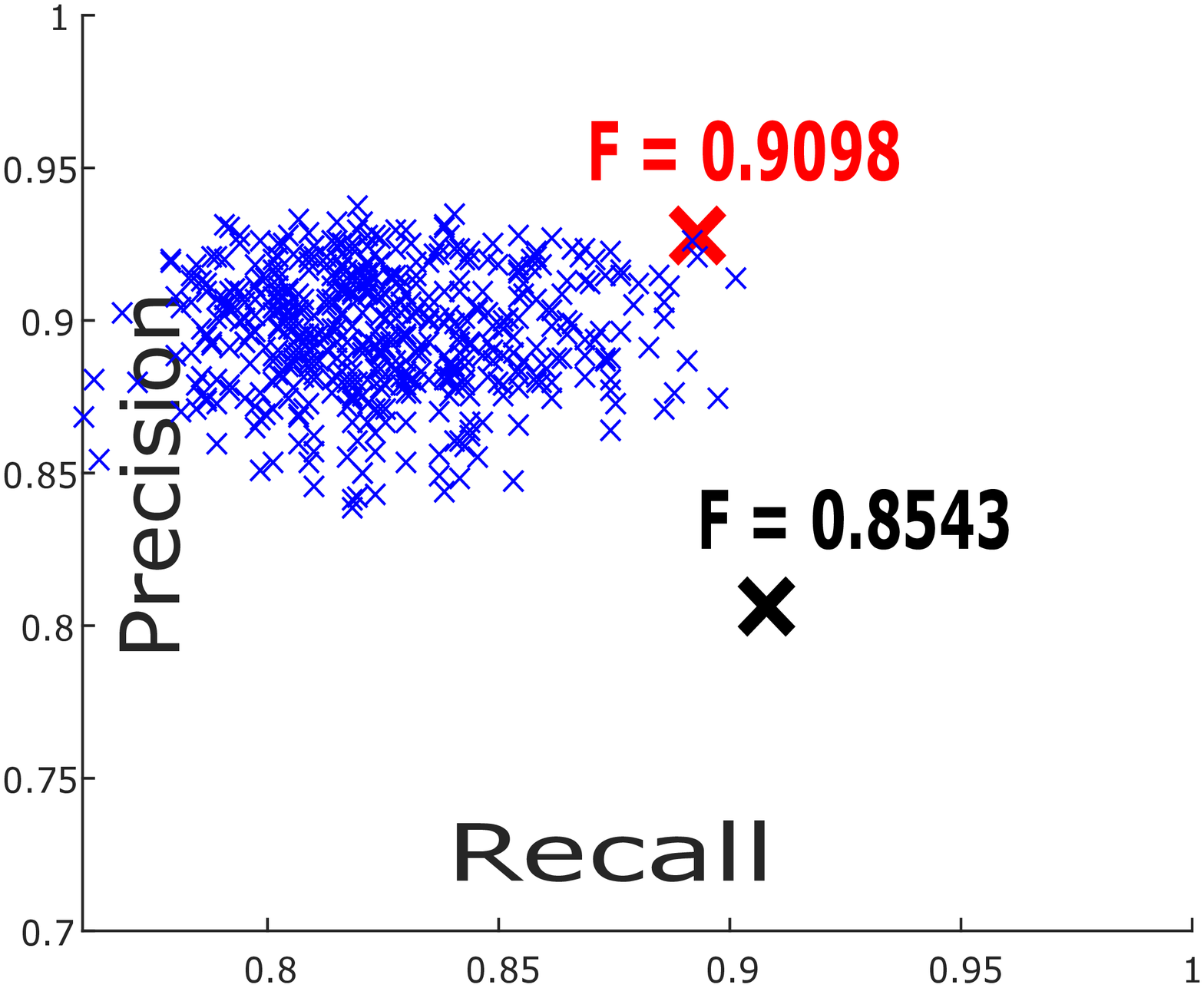}                                                                    }
\end{tabular}
 \caption{Experiments on walking silhouettes data set. In the PR curves, the \lq\textcolor{red}{$\times$}\rq marks the sample having the best F-measure value obtained using the proposed approach (with either global or local shape priors), and the \lq$\times$\rq marks that of segmentation of Kim et al.~\cite{kim2007nonparametric}.\label{fig:experiments_walking}}
\end{figure*}

\section{Conclusion}
We have presented a MCMC shape sampling approach for image segmentation that exploits prior information about the shape to be segmented. Unlike existing MCMC sampling methods for image segmentation, our approach can segment objects with occlusion and suffering from severe noise, using nonparametric shape priors. We also provide an extension of our method for segmenting shapes of objects with parts that can go through independent shape variations by using local shape priors on object parts. Empirical results on various data sets demonstrate the potential of our approach in MCMC shape sampling. The implementation of the proposed method is available at \url{spis.sabanciuniv.edu/data_code}.

\vfill
\clearpage
\newpage
{
\bibliographystyle{ieee}
\bibliography{egbib}

\begin{thebibliography}{10}\itemsep=-1pt

\bibitem{chan2001active}
T.~F. Chan, L.~Vese, et~al.
\newblock Active contours without edges.
\newblock {\em IEEE transactions on Image processing}, 10(2):266--277, 2001.

\bibitem{chang2011efficient}
J.~Chang and J.~Fisher.
\newblock Efficient mcmc sampling with implicit shape representations.
\newblock In {\em IEEE Conference on Computer Vision and Pattern Recognition
  (CVPR)}, pages 2081--2088. IEEE, 2011.

\bibitem{chang2012efficient}
J.~Chang and J.~W. Fisher~III.
\newblock Efficient topology-controlled sampling of implicit shapes.
\newblock In {\em Proceedings of the IEEE International Conference on Image
  Processing (ICIP)}, Sept 2012.

\bibitem{chen2013deep}
F.~Chen, H.~Yu, R.~Hu, and X.~Zeng.
\newblock Deep learning shape priors for object segmentation.
\newblock In {\em IEEE Conference on Computer Vision and Pattern Recognition
  (CVPR)}, pages 1870--1877. IEEE, 2013.

\bibitem{chen2009markov}
S.~Chen and R.~J. Radke.
\newblock Markov chain monte carlo shape sampling using level sets.
\newblock In {\em IEEE 12th International Conference on Computer Vision
  Workshops (ICCV Workshops)}, pages 296--303. IEEE, 2009.

\bibitem{cremers2006kernel}
D.~Cremers, S.~J. Osher, and S.~Soatto.
\newblock Kernel density estimation and intrinsic alignment for shape priors in
  level set segmentation.
\newblock {\em International Journal of Computer Vision}, 69(3):335--351, 2006.

\bibitem{de2004image}
M.~De~Bruijne and M.~Nielsen.
\newblock Image segmentation by shape particle filtering.
\newblock In {\em Proceedings of the 17th International Conference on Pattern
  Recognition (ICPR)}, volume~3, pages 722--725. IEEE, 2004.

\bibitem{eslami2014shape}
S.~A. Eslami, N.~Heess, C.~K. Williams, and J.~Winn.
\newblock The shape boltzmann machine: a strong model of object shape.
\newblock {\em International Journal of Computer Vision}, 107(2):155--176,
  2014.

\bibitem{fan2007mcmc}
A.~C. Fan, J.~W. Fisher~III, W.~M. Wells~III, J.~J. Levitt, and A.~S. Willsky.
\newblock Mcmc curve sampling for image segmentation.
\newblock In {\em Medical Image Computing and Computer-Assisted Intervention
  (MICCAI)}, pages 477--485. Springer, 2007.

\bibitem{gilks1996}
W.~R. Gilks, S.~Richardson, and D.~J. Spiegelhalter.
\newblock (1996). markov chain monte carlo in practice.

\bibitem{houhou2008fast}
N.~Houhou, J.-P. Thiran, and X.~Bresson.
\newblock Fast texture segmentation model based on the shape operator and
  active contour.
\newblock In {\em IEEE Conference on Computer Vision and Pattern Recognition
  (CVPR)}, pages 1--8. IEEE, 2008.

\bibitem{kim2007nonparametric}
J.~Kim, M.~{\c{C}}etin, and A.~S. Willsky.
\newblock Nonparametric shape priors for active contour-based image
  segmentation.
\newblock {\em Signal Processing}, 87(12):3021--3044, 2007.

\bibitem{kim2005nonparametric}
J.~Kim, J.~W. Fisher~III, A.~Yezzi, M.~{\c{C}}etin, and A.~S. Willsky.
\newblock A nonparametric statistical method for image segmentation using
  information theory and curve evolution.
\newblock {\em IEEE Transactions on Image Processing}, 14(10):1486--1502, 2005.

\bibitem{lecun1998gradient}
Y.~LeCun, L.~Bottou, Y.~Bengio, and P.~Haffner.
\newblock Gradient-based learning applied to document recognition.
\newblock {\em Proceedings of the IEEE}, 86(11):2278--2324, 1998.

\bibitem{mesadi2015disjunctive}
F.~Mesadi, M.~Cetin, and T.~Tasdizen.
\newblock Disjunctive normal shape and appearance priors with applications to
  image segmentation.
\newblock In {\em Medical Image Computing and Computer-Assisted Intervention
  (MICCAI)}, pages 703--710. Springer, 2015.

\bibitem{metropolis1953equation}
N.~Metropolis, A.~W. Rosenbluth, M.~N. Rosenbluth, A.~H. Teller, and E.~Teller.
\newblock Equation of state calculations by fast computing machines.
\newblock {\em The journal of chemical physics}, 21(6):1087--1092, 1953.

\bibitem{michailovich2007image}
O.~Michailovich, Y.~Rathi, and A.~Tannenbaum.
\newblock Image segmentation using active contours driven by the bhattacharyya
  gradient flow.
\newblock {\em IEEE Transactions on Image Processing}, 16(11):2787--2801, 2007.

\bibitem{mumford1989optimal}
D.~Mumford and J.~Shah.
\newblock Optimal approximations by piecewise smooth functions and associated
  variational problems.
\newblock {\em Communications on pure and applied mathematics}, 42(5):577--685,
  1989.

\bibitem{salakhutdinov2009deep}
R.~Salakhutdinov and G.~E. Hinton.
\newblock Deep boltzmann machines.
\newblock In {\em International conference on artificial intelligence and
  statistics}, pages 448--455, 2009.

\bibitem{tsai2003shape}
A.~Tsai, A.~Yezzi~Jr, W.~Wells, C.~Tempany, D.~Tucker, A.~Fan, W.~E. Grimson,
  and A.~Willsky.
\newblock A shape-based approach to the segmentation of medical imagery using
  level sets.
\newblock {\em IEEE Transactions on Medical Imaging}, 22(2):137--154, 2003.

\end{thebibliography}
}

\end{document}